\pdfoutput=1
\documentclass[11pt]{article}

\usepackage[final]{acl}

\usepackage{times}
\usepackage{latexsym}

\usepackage[T1]{fontenc}
\usepackage[utf8]{inputenc}

\usepackage{microtype}

\usepackage{inconsolata}

\usepackage{multirow}
\usepackage{graphicx}
\usepackage{pifont}
\usepackage{booktabs}
\usepackage{subfigure}
\usepackage{algorithm}
\usepackage{algorithmic}
\usepackage{amsfonts}
\usepackage{amsmath}
\usepackage{amssymb}
\usepackage{mathrsfs}
\usepackage{xcolor,colortbl}
\usepackage{enumitem} 
\usepackage{arydshln}
\usepackage{multicol}
\usepackage{caption}
\usepackage{subcaption}
\usepackage{tcolorbox}
\usepackage{mdframed}
\usepackage{float}

\definecolor{CommentGrey}{HTML}{949494}
\definecolor{DeepForest}{HTML}{227805}
\definecolor{DeepBlood}{HTML}{780505}
\definecolor{Twitch}{HTML}{8c34eb}
\definecolor{Saffron}{HTML}{eb5c34}
\definecolor{Navy}{HTML}{2c59a5}

\newtcbox{\mybox}[1][red]
  {on line, arc = 0pt, outer arc = 0pt,
    colback = #1!10!white, colframe = #1!50!black,
    boxsep = 0pt, left = 1pt, right = 1pt, top = 2pt, bottom = 2pt,
    boxrule = 0pt, bottomrule = 1pt, toprule = 1pt}

\newtcolorbox{mydefinition}[1]{colback=gray!10!white, colframe=gray!75!black,
    fonttitle=\bfseries, title=#1}

\definecolor{Gray}{gray}{0.88}
\definecolor{LightCyan}{rgb}{0.88,1,1}

\title{Tokenization Falling Short: On Subword Robustness in Large Language Models}

 \author{Yekun Chai$^{*\spadesuit}$\, Yewei Fang\thanks{Equal contribution and shared co-first authorship.}$^{\heartsuit}$\, Qiwei Peng$^\diamondsuit$ Xuhong Li$^\spadesuit$ \\
  $^\spadesuit$Baidu \, 
  $^\heartsuit$ModelBest \, 
  $^\diamondsuit$University of Copenhagen \\
  \texttt{\{chaiyekun,ywfcast1e\}@gmail.com}\\ 
} 
\begin{document}
\maketitle
\begin{abstract}
Language models typically tokenize raw text into sequences of subword identifiers from a predefined vocabulary, a process inherently sensitive to typographical errors, length variations, and largely oblivious to the internal structure of tokens—issues we term \mybox[blue]{\textit{the curse of tokenization}}.  In this study, we delve into these drawbacks and demonstrate that large language models (LLMs) remain susceptible to these problems. This study systematically investigates these challenges and their impact on LLMs through three critical research questions: (1) complex problem solving, (2) token structure probing, and (3) resilience to typographical variation. Our findings reveal that scaling model parameters can mitigate the issue of tokenization; however, LLMs still suffer from biases induced by typos and other text format variations. Our experiments show that subword regularization such as BPE-dropout can mitigate this issue. We release our evaluation code and data at \url{https://github.com/FloatAI/TKEval}.

\end{abstract} 

\section{Introduction}
\label{sec:intro}

Tokenization is a fundamental step in the preprocessing pipeline of large language models (LLMs) \cite{gpt4, gemini23,llama2,ernie-code,starcoder2-24}, converting raw text into a sequence of subword units derived from a predefined vocabulary~\cite{bpe,spm18}. This process, while effective in many scenarios, presents significant challenges that can hinder the performance and robustness of LLMs. These challenges include sensitivity to typographical errors~\cite{CaoKMI23}, length variations~\cite{htlm22}, and a lack of awareness of the internal structure of tokens~\cite{brown2020language}—collectively termed \textit{the curse of tokenization}.

\begin{figure}[!ht] 
\centering
% \vspace{-1em}
\subfigure[cosine (\texttt{``assignment''}, \newline \texttt{``assign'' + ``ment''}).] { \label{fig:vec_assignment}
\includegraphics[width=0.46\linewidth]{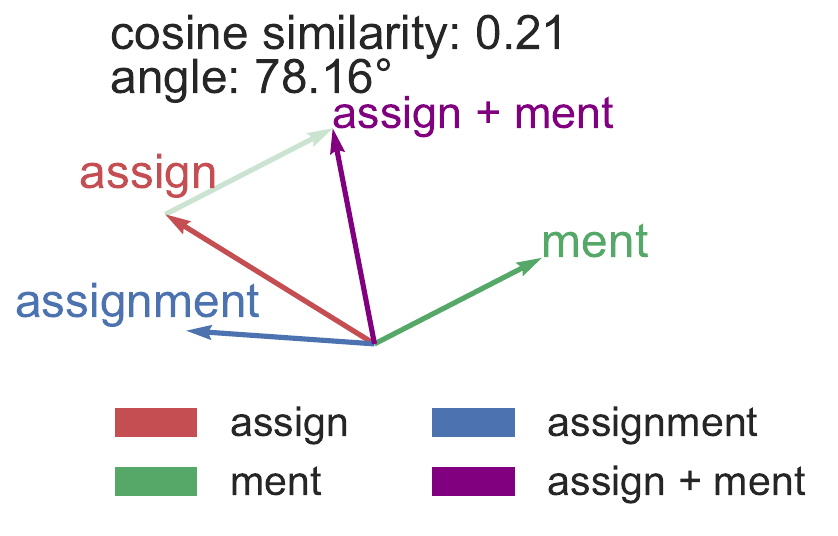}
}
\hfill
\subfigure[cosine(\texttt{``import''}, \newline   \texttt{``im''} + \texttt{``port''}).] { \label{fig:vec_import}
\includegraphics[width=0.46\linewidth]{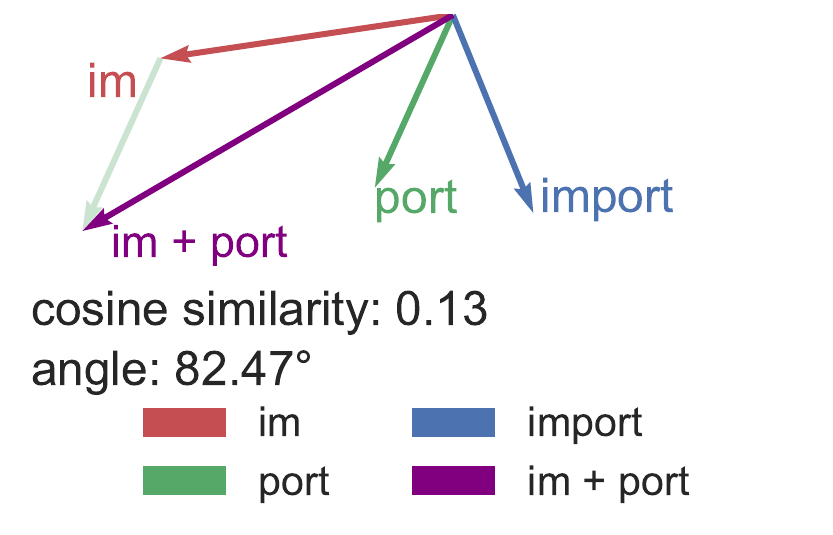}
}
\vspace{-8pt}
\caption{Compositional challenges in token embeddings. (a) \texttt{``assignment''} decomposed into \texttt{``assign''} and \texttt{``ment''} shows a cosine similarity of 0.21 and an angle of 78.16°. (b) \texttt{``import''} decomposed into \texttt{``im''} and \texttt{``port''} shows a cosine similarity of 0.13 and an angle of 82.47°. These results indicate that existing LLMs do not accurately capture surface form composition.
}
\label{fig:vec_plot}
\vspace{-20pt}
\end{figure}

Typographical errors, such as minor misspellings or misplaced characters, can drastically affect the tokenization process. Unlike humans, who can easily overlook these errors and understand the intended meaning, LLMs can misinterpret or fail to recognize these variations, leading to degraded performance~\cite{CaoKMI23}. This typo-sensitivity reveals a crucial gap in current tokenization methods, which do not sufficiently mimic human reading capabilities.

Another critical issue is the length unawareness of current tokenization approaches. LLMs often struggle to accurately represent the organizational structure of text, being insensitive to the number of characters or words~\cite{htlm22}. This insensitivity affects their ability to understand and process text effectively, particularly in tasks requiring a nuanced understanding of text length and compositional structure.

Furthermore, existing tokenization methods are often blind to the internal structure of tokens. The decoupled embedding space and lookup table approach \textbf{fail to account for the hierarchical composition of language}, spanning characters, subwords, and words, as depicted in Figure~\ref{fig:vec_plot}. This lack of integration across different levels of token composition limits the model's ability to fully grasp semantic relationships and differences. 

Case insensitivity in certain languages adds another layer of complexity, where variations in capitalization can lead to different token representations, further complicating the model's understanding and processing of text.

% To address the challenges posed by current tokenization methods, we conducted a comprehensive study examining their limitations and the resulting impact on LLM performance. Our investigation focuses on three key areas:

% First, we explore \textbf{complex problem solving}, investigating how LLMs handle tasks that are sensitive to tokenization, such as anagram solving and complex mathematical language understanding  (\S\ref{sec:rq1}). These tasks allow us to assess the model's ability to navigate scenarios where tokenization plays a crucial role in performance.

% Second, we delve into \textbf{token structure probing} (\S\ref{sec:rq2}), studying tasks related to token-level understanding, such as case manipulation, length counting, and length-sensitive operations. This analysis helps us evaluate how well LLMs comprehend and process token structures in various contexts.

% Third, we evaluate the resilience of LLMs to \textbf{typographical variation} (\S\ref{sec:rq3}). We design robust evaluation benchmarks based on datasets like MMLU, TruthfulQA, GSM8K, and HumanEval, covering diverse tasks and linguistic phenomena to systematically test the impact of typographical errors and text format inconsistencies on model performance.

To address these challenges, we conducted a comprehensive study examining the limitations of current tokenization methods and their impact on LLM performance. Our study is guided by three critical research questions:

\vspace{0.1em}\noindent \textbf{A. Complex Problem Solving} (\S\ref{sec:rq1}). As a pilot experiment, we firstly investigate the performance of LLMs on complex problems that are sensitive to tokenization, involving anagram task and complex mathematical language understanding.
% rearranging or unscrambling letters

\vspace{0.1em}\noindent \textbf{B. Token Structure Probing} (\S\ref{sec:rq2}). We study the token structural tasks such as case manipulation, length counting, and length-sensitive tasks to probe the token structural understanding of LLMs.

\vspace{0.1em}\noindent \textbf{C. Typographical Variation} (\S\ref{sec:rq3}). We designed a robust set of evaluation benchmarks on top of various datasets such as MMLU, TruthfulQA, GSM8K, and HumanEval, covering diverse tasks and linguistic phenomena. These benchmarks allow us to systematically test and analyze the LLMs' resilience to tokenization.

Our findings highlight that while scaling model parameters can enhance the robustness to tokenization, LLMs still suffer from biases introduced by typographical errors and text format variations. We demonstrate the persistent nature of these tokenization challenges.

\paragraph{Contribution} To conclude, our main contributions are as follows:
\begin{enumerate}[noitemsep, left=0pt, labelsep=4pt,]
    \item We provide a comprehensive analysis of the problem known as the \textit{curse of tokenization}, detailing its impact on language model performance and introducing systematic evaluation benchmarks to assess these issues.
    \item By evaluating various scales of LLMs, including LLama3, Mistral, and GPT-4 families, across thirteen distinct tasks, we demonstrate that even state-of-the-art models struggle with handling typographical variations. Specifically, LLMs exhibit greater sensitivity to character-level variations compared to subword-level variations.
    \item We demonstrate that regularized tokenization approaches, such as BPE-dropout with moderate drop rates, can enhance the model's resilience to the discussed issues.
    % \item To facilitate further research, we release our evaluation code and benchmarks, enabling the research community to build upon our findings and develop more robust models.
\end{enumerate} 

% \noindent\textbf{2)} 

% This evaluation aims to provide a foundation for future research aimed at developing more resilient tokenization strategies.

% \noindent\textbf{3)} 

% \noindent\textbf{4)} 

% We provide a comprehensive analysis of the problem of \textit{curse of tokenization} and introduces evaluation benchmarks to systematically assess these issues. By evaluating across various LLM scales of LLama3, Mistral and GPT-4 families on thirteen tasks, we aim to shed light on the critical areas needing improvement and provide a foundation for future research aimed at developing more resilient tokenization strategies. 
% Our results show that even the state-of-the-art models struggle with handling the typographical variations. In particular, LLMs are much more sensitive to character-level variations in comparison to subword-level variations. We will release our evaluation code and benchmarks to facilitate further research.

% \cyk{TBA: more concl}

\section{Related Work}
\label{sec:bg}

\subsection{Tokenization}
\paragraph{Tokenization Approach} Conventional language models~\cite{radford2018improving, jozefowicz2016exploring, brown2020language} typically tokenize input text into a sequence of tokens by splitting it into smaller subwords. Traditional tokenization approaches include frequency-based methods such as Byte Pair Encoding (BPE; \citealp{bpe}) and probability-based methods like WordPiece~\cite{DBLP:conf/icassp/SchusterN12}. BPE merges tokens based on bigram frequency, relying on subword pair co-occurrence to greedily merge neighboring pairs. In contrast, WordPiece can be viewed as a language-modeling based BPE variant. It select the unit pair that maximizes the bigram likelihood of training data at utmost, rather than choose the most frequent pair. 

% This approach uses a greedy longest-match-first algorithm, which does not generate multiple segmentations with probabilities.

Unigram Language Model~\cite{kudo-2018-subword} prunes tokens based on unigram LM perplexity, treating the segmentation process as a probabilistic mixture of characters, subwords, and words, reducing subwords by evaluating likelihood reduction. Additionally, some tokenization methods handle text at the byte level~\cite{byt5} or character level~\cite{10.5555/3104482.3104610, clark-etal-2022-canine}. Conventional LLMs often use byte-level BPE (BBPE) for base vocabulary construction, representing any text with a moderate vocabulary size and avoiding the out-of-vocabulary (OOV) problem. For a detailed introduction to tokenization, readers can refer to \citet{chai2021tokenization-PTMs}.

\paragraph{Tokenization-Free Approach} Tokenization approaches often suffer from the \textit{vocabulary bottleneck}, where there is a trade-off between vocabulary size and  diverse language coverage in multilingual scenarios. To address this issue, \citet{pixel23} and \citet{chai2024autoregressive} introduced a tokenization-free approach that renders raw text as visual text images for masked language modeling and autoregressive pre-training. This method demonstrates robust multilingual generalization capabilities compared to subword tokenization approaches.

\subsection{Perturbation Probing}
\label{sec:anagram}
Several studies have investigated the behavior of language models under input perturbations at various levels, including character-level~\cite{nishino-etal-2019-generating}, subword-level~\cite{abdou-etal-2022-word}, and word-level~\cite{sinha-etal-2021-unnatural,peng-etal-2023-testing} scrambling. Despite these efforts, intrinsic evaluations of perturbing LLM inputs remain under-explored.

\citet{CaoKMI23} proposed examining scrambled sentence recovery and scrambled QA with context corruptions. In contrast, our study conducts a comprehensive evaluation of both character- and subword-level perturbations, along with noise injection. We evaluate a wide range of LLMs across various tasks to provide a detailed comparison and inspire future research in tokenization and robust model performance.

% Our work expands on previous studies by providing a thorough analysis of how different types of perturbations affect mainstream LLMs. This includes examining the resilience of these models to various forms of input noise and scrambling, offering insights that are crucial for developing more robust tokenization strategies.

 % anagram task bg

% abdou-etal-2022-word,
%  % word-level / token-level
% } % char-level

% RQ1: 
% Blind to Token Internals/Composition
% Length-Unaware, Counting Challenges
% Case-Insensitive

\section{Complex Problem Solving}
\label{sec:rq1}
Complex problem-solving tasks are critical benchmarks for evaluating the complex reasoning and comprehension capabilities of LLMs. We explore the LLM's ability to perform intricate operations on tokenized inputs, as the tokenization process is fundamental to determine how the raw text is segmented and processed, directly impacting the model's interpretation and prediction.

Anagram solving and mathematical language comprehension were selected to elucidate the relationship between tokenization quality and the model's performance on complex problem-solving. {Anagram tasks} require models to decode and rearrange jumbled letters into coherent words, emphasizing the importance of precise token boundaries and recognition accuracy. On the other hand, {mathematical language comprehension}, particularly expressed with {\LaTeX}-formatted expressions, demands an exact interpretation of specialized symbols and structured notation, challenging the tokenization process's robustness.

\subsection{Anagram Task}
% Surface-Form and Structural Composition

% \footnote{\url{https://github.com/google/BIG-bench/tree/main/bigbench/benchmark_tasks/cycled_letters}}
% \footnote{\url{https://github.com/google/BIG-bench/blob/main/bigbench/benchmark_tasks/word_unscrambling/}}

\paragraph{Task Description and Settings} The anagram task tests the model's ability to unscramble a sequence of jumbled characters to form a valid word. This task evaluates the model's handling of surface-form manipulations and its understanding of char-level compositions within a word. The complexity arises from the need to identify potential word candidates from mixed characters and reassemble them correctly. We present a task example in \S\ref{ap:example-complex}. Specifically, we include two tasks:

\begin{figure}[!ht]
    % \vspace{-1mm}
    \centering
    \includegraphics[width=0.75\linewidth]{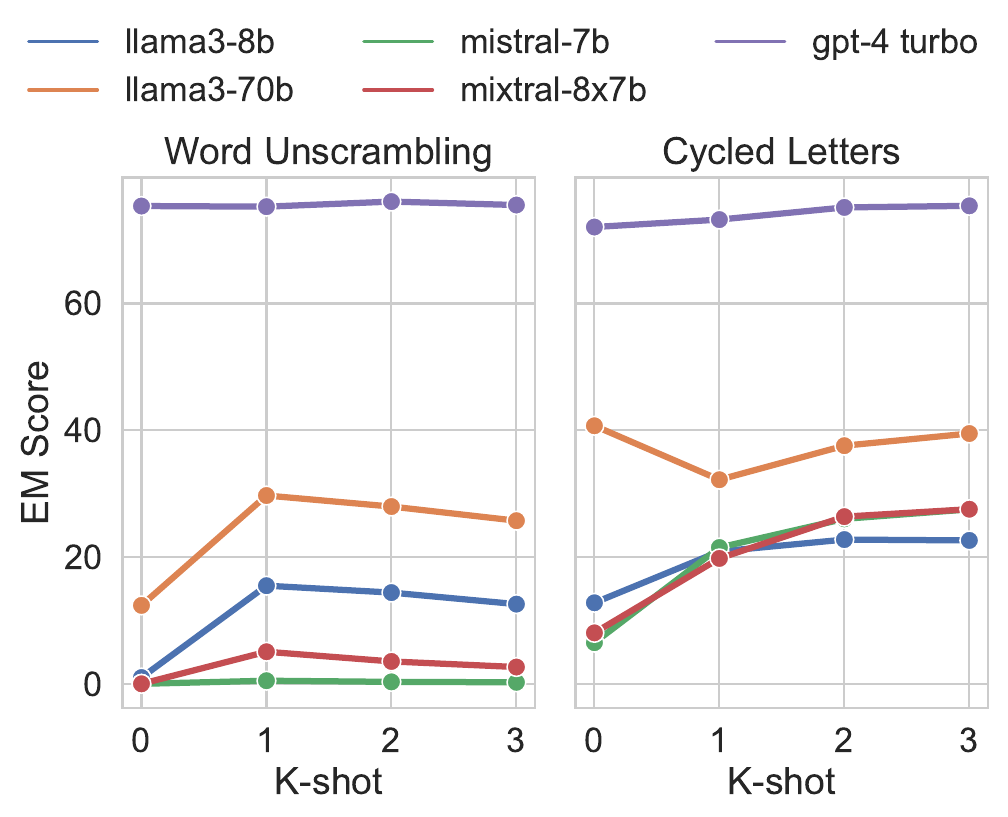}
     % \vspace{-8pt}
    \caption{$K$-shot performance on Word Unscrambling (WU) and Cycled Letters (CL) tasks. The plots illustrate that increasing the number of demonstration examples ($K$-shot) does not consistently enhance performance. However, models with larger parameter sizes generally exhibit better performance across both tasks.}
    \label{fig:anagram-plot} 
    % \vspace{-10pt}
\end{figure}
% \comm{add results line plot here, see Figure 3.11 in GPT-3 paper}
% topsep=0pt, partopsep=0pt
\begin{itemize}[noitemsep, left=0pt, labelsep=4pt,]
    \item \textbf{Cycled Letters in Word (CL; \citealp{bigbench22})} -- The model is given a word with its letters cycled, and is expected to generate the original word (\emph{e.g.}, ``remo'' $\rightarrow$ ``more'').
    \item \textbf{Word Unscrambling (WU; \citealp{bigbench22})} -- The model is given a randomly scrambled word, and must recover the original word (\emph{e.g.}, ``nad'' $\rightarrow$ ``and'').
\end{itemize}

\noindent We employ exact match (EM) scores for evaluation. Unless otherwise specified, we use the inference-time temperature of $0$ for all LLMs in following experiments, to assure the results reproducible.

\begin{figure}[!ht]
    % \vspace{-10pt}
    \centering
    \includegraphics[width=0.65\linewidth]{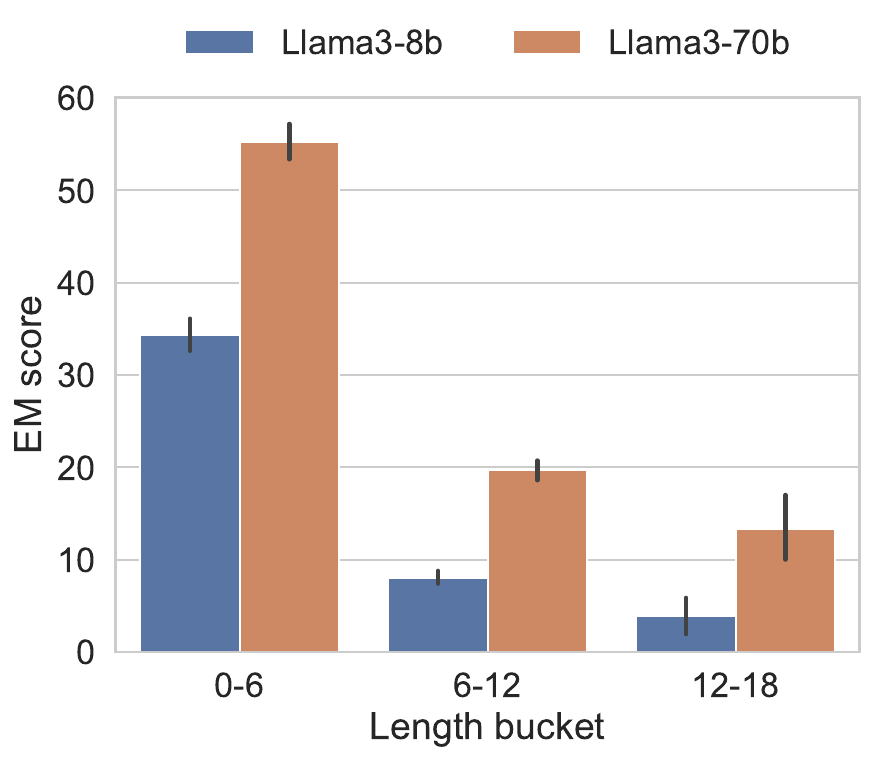}
     \vspace{-1em}
    \caption{The relationship between the length of scrambled words and the Exact Match (EM) score of Llama3-8B and Llama3-70B on the word unscrambling task under one-shot evaluation. The models tend to correctly reorder anagrams of shorter lengths, while struggling with longer words.}
    \label{fig:WU-boxplot-llama3} 
    % \vspace{-10pt}
\end{figure}

\paragraph{Results and Analysis} The experimental results reveal that larger models demonstrate better performance on the anagram task, yet they remain susceptible to tokenization errors. Specifically, models struggled with longer anagrams (see Figure~\ref{fig:WU-boxplot-llama3}) or those containing uncommon letter combinations, indicating that while scaling improves token recognition, inherent tokenization flaws persist. 

Figure~\ref{fig:WU-boxplot-llama3} highlights the performance differences between Llama3-8B and Llama3-70B on the word unscrambling task across various word lengths. Notably, Llama3-70B consistently outperforms Llama3-8B, especially in the 0-6 and 6-12 character buckets. This trend indicates that as the model parameter size increases from 8B to 70B, the ability to accurately reorder scrambled words improves. However, both models struggle with longer words (12-18 characters), though Llama3-70B maintains a moderate edge. 

% \comm{+ box plot.=> len(word) vs. EM score}

Our results shown in Figure~\ref{fig:anagram-plot} indicate several key trends. Firstly, the performance improves significantly as model size scales from 8B to 70B parameters~\cite{llama3modelcard}. Secondly, we observe that while the dense Mistral-7B model performs poorly, the sparse Mixtral-8x7B model (an MoE sparse model) shows improved performance due to its parameter size scaling. Lastly, GPT-4 turbo, a much more powerful model, achieves state-of-the-art results, clearly outperforming all other models across all shot conditions.
This sensitivity underscores the need for more robust tokenization that can handle typographical variations without degrading performance.

\subsection{Mathematical Language ({\LaTeX}) Comprehension}
\noindent \textbf{Task Description and Settings}  The mathematical language comprehension task evaluates the model's ability to read and comprehend mathematics written in {\LaTeX}, the typesetting language used by professional mathematicians. This task assesses the models' capability to interpret complex mathematical expressions and accurately tokenize symbols and operators within structured {\LaTeX} format. 

We employ \textbf{Identify Math Theorems (IMT; \citealp{bigbench22})} for evaluation, and use perplexity to measure the model's confidence in different given choices. The dataset comprises 54 problems divided into nine sections, each representing a major area of mathematics research or advanced pedagogy. We present the input-output format in \S\ref{ap:example-complex}.

\begin{table}[!ht]
\centering
\resizebox{\columnwidth}{!}{
\ttfamily
\begin{tabular}{@{}lllll@{}}
\toprule
Setting       & \textbf{0-Shot} & \textbf{1-Shot} & \textbf{2-Shot} & \textbf{3-Shot} \\ \midrule
GPT-3 (6B)$^a$     & 33.96           & 28.30           & 33.96           & 28.30           \\
GPT-3 (200B)$^a$   & 32.08           & 30.19           & 33.96           & 30.19           \\
Llama2-7b    & 37.70           & 34.00           & 35.80           & 37.70           \\
Llama3-8b    & 41.51           & 45.28           & 45.28           & 35.85           \\
Llama3-70b   & \textbf{62.26}  & \textbf{79.25}  & \textbf{69.81}  & \textbf{71.70}  \\
Mistral-7b   & 47.20           & 43.40           & 37.70           & 37.70           \\
Mixtral-8x7b & 49.10           & 56.60           & 64.20           & 62.30           \\ \bottomrule
\end{tabular}%
}
\caption{Few-shot results on \textbf{Identifying Math Theorems}, with exact match scores reported as percentages. $^a$ refers to results taken from \citet{bigbench22}}
\label{tab:rq1-math}
\vspace{-1em}
\end{table}

\paragraph{Results and Analysis}  Our evaluation results of various models are shown in Table~\ref{tab:rq1-math}. The results demonstrate that while larger models generally perform better on LaTeX-formatted mathematical content, the relationship between the number of in-context examples and model performance is not linear. The Llama3-70B model consistently outperformed other models, achieving a score of \texttt{62.26\%} in the zero-shot setting and improving to \texttt{79.25\%} with one-shot learning. However, additional in-context examples led to fluctuating performance, with scores of \texttt{69.81\%} in the two-shot setting and \texttt{71.70\%} in the three-shot setting. This indicates that increasing the number of in-context demonstrations does not consistently enhance performance and may lead to variability depending on the specific examples provided.

Other models exhibited similar trends. GPT-3-200B~\cite{bigbench22}, despite its larger parameter count, did not show significant improvement over the smaller GPT-3-6B model, suggesting that simply increasing the model size does not guarantee better performance in {\LaTeX} comprehension tasks. The comparison between dense and sparse models revealed that the dense Mistral-7B model performed poorly across all shot conditions, while the sparse Mixtral-8x7B model demonstrated better results, especially in few-shot scenarios. 

% Sum up
% These observations highlight several key points. (1) Increasing the number of in-context examples does not necessarily improve model performance, as evidenced by the fluctuating results of Llama3-70B. This suggests that models may not fully comprehend additional context after tokenization for such complex tasks. (2) Scaling model parameters generally leads to better performance, with Llama-70B significantly outperforming smaller models. (3) Sparse models like Mixtral-8x7B offer advantages over dense models, particularly in handling few-shot learning scenerios.

\section{Token Structure Probe}
\label{sec:rq2}
Tokenization is a key preprocessing step in LLMs, yet it introduces several significant challenges, which we defined as \mybox[blue]{\textit{the curse of tokenization}}. These challenges include length unawareness, case insensitivity, and a lack of awareness of the internal structure of tokens. Tokenization transforms text into sequences of token identifiers, often obscuring the surface form and internal structure of the original text. This conversion can lead to deficiencies in the model's ability to understand and process textual data accurately.

 % topsep=0pt, partopsep=0pt
 
\textit{The curse of tokenization} manifests in several ways, which refers to the inherent challenges:
\begin{enumerate}[label=\Alph*), noitemsep, left=0pt, labelsep=4pt,]
\item \textbf{Length Unawareness}: Models struggle to recognize the organizational structure of text, such as the number of characters or words.
\item \textbf{Case Insensitivity}: Variations in capitalization can lead to different token identifiers and representations, complicating the model's processing of text.
\item \textbf{Blindness to Internal Structure}: The decoupled embedding space and lookup table approach used in LLMs fail to preserve the hierarchical and relational structure within tokens, obscuring the surface form and internal relationships between characters and subwords.
\end{enumerate}

To address these challenges, we construct a set of probing tasks to evaluate the model's understanding of token structure. These tasks are divided into intra-token (\S\ref{sec:intra-token}) and inter-token probes (\S\ref{sec:inter-token}).

\subsection{Intra-Token Probing}
\label{sec:intra-token}

\paragraph{Task Description and Settings}To measure the capability of LLMs, we devise intra-token probing tasks related to length, case, and counting problems. These tasks evaluate the model's performance on the internal structure of tokens or word, specifically including four tasks:

% topsep=0pt, partopsep=0pt
\begin{itemize}[noitemsep, left=0pt, labelsep=4pt, ]
    \item \textbf{Character Count (CC)} -- The model is asked to count the number of occurrences of a specific character within the given word (\emph{e.g.}, the character appears twice in the word ``undertake'' $\rightarrow$ the answer is: ``e'').
    \item \textbf{$N$-th Character (NC)} -- The model is expected to output the $n$-th character of the given word (\emph{e.g.}, 4-th character of the word ``dual'' $\rightarrow$ ``l'').
    \item \textbf{$N$-th Character Reverse (NCR)} -- The model must identify the $n$-th character from the end of a word (\emph{e.g.}, 2nd character from the end of the word ``dual'' $\rightarrow$ ``a'').
    \item \textbf{Case Conversion (CCV)} -- This task involves converting the characters within a word to different cases (uppercase, lowercase) or converting the word into title case.
\end{itemize}

For each task, we conduct many-shot evaluation (0-3 shot) and report the EM score to test the model's ability to understand and manipulate the internal structure of tokens and words at a granular level, revealing the extent to which the tokenization process could preserve this information. Detailed test examples are provided in Appendix~\ref{ap:example-intra-token}.

\begin{figure}[!ht]
    % \vspace{-1mm}
    \centering
    \includegraphics[width=0.95\linewidth]{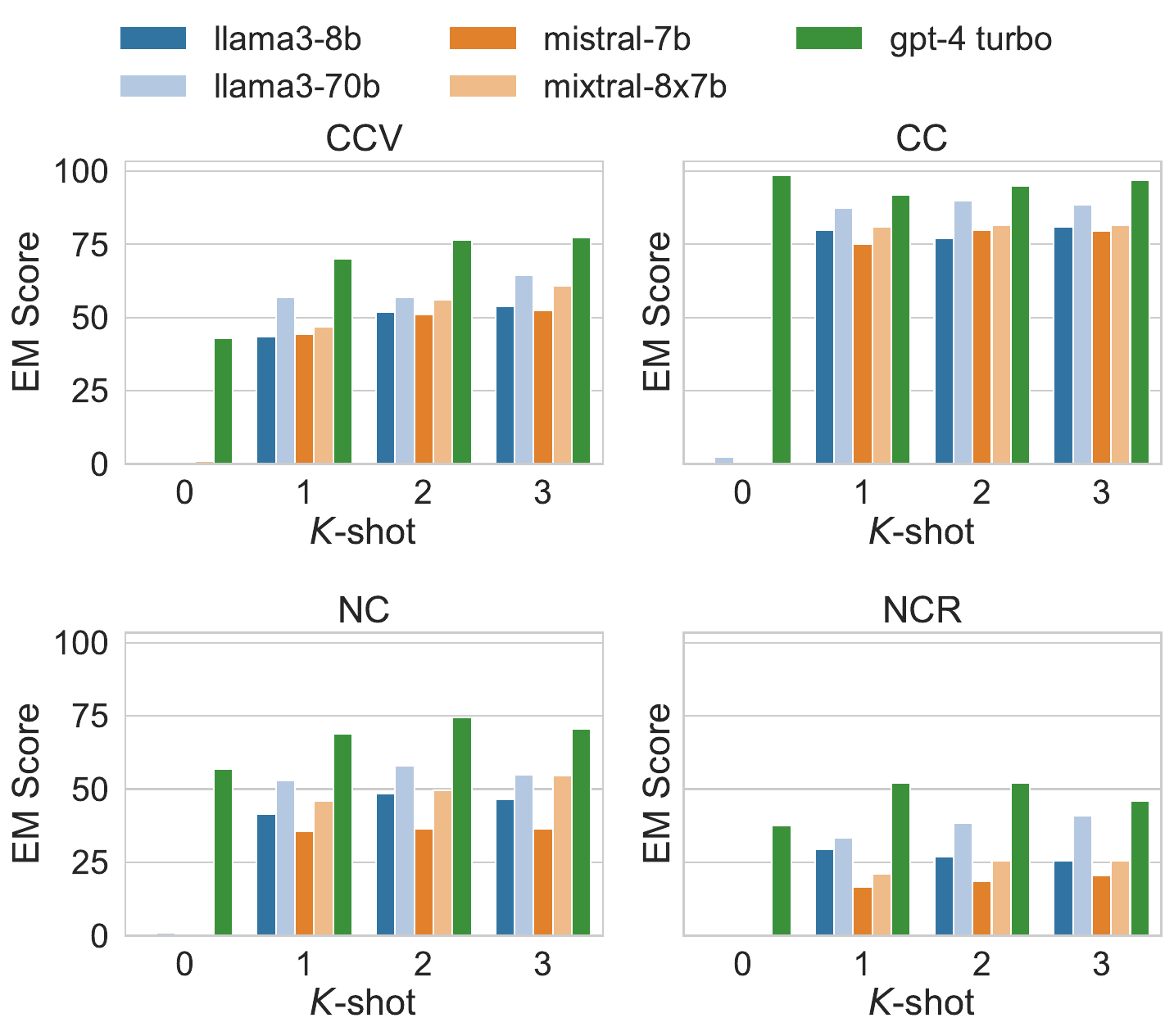}
     % \vspace{-1em}
    \caption{$K$-shot performance on intra-token probing tasks (CCV, CC, NC, NCR). The plots demonstrate that increasing the number of demonstration examples ($K$-shot) generally results in an improvement from zero-shot to one-shot, with performance stabilizing thereafter. }
    \label{fig:intra-token-plot} 
    \vspace{-10pt}
\end{figure}

% \comm{add results table or barplot here}

\paragraph{Results and Analysis} We evaluated CC, NC, NCR, and CCV tasks across different models and shot settings. The results are presented in Figure~\ref{fig:intra-token-plot}. 

The CC task reveals that larger models exhibit superior performance, particularly GPT-4 turbo, which achieves near-perfect accuracy across all shot conditions. Smaller models, such as Llama3-8B, show significant improvement with few-shot learning, indicating that exposure to examples greatly enhances their performance. For example, Llama3-8B's accuracy improves from 0\% in the zero-shot setting to 81\% in the three-shot setting. This demonstrates that increased model size and few-shot learning contribute positively to CC tasks.

The NC task underscores the difficulty models face in accurately identifying specific characters within words. GPT-4 turbo again leads in performance, while smaller models show substantial improvement with increased shots. Llama3-70B, for instance, improves from 1\% in the zero-shot setting to 55\% in the three-shot setting. This indicates that while larger models perform better, few-shot learning plays a crucial role in enhancing the model's ability to identify specific characters.

Identifying characters from the end of the word, or reverse character identification, proves more challenging. GPT-4 turbo achieves the highest performance with a score of 52\% in the one-shot setting, though overall accuracy is lower compared to other tasks. Smaller models like Llama3-8B show moderate improvements with additional shots, but their performance remains relatively low. This highlights the complexity of reverse character identification and the need for more advanced tokenization strategies to address this challenge.

% CC tasks, which involve changing characters within a word to different cases, are particularly challenging. All models show improvement with increased shots, with GPT-4 turbo consistently leading in performance. Notably, Llama3-70B and Mixtral-8x7B demonstrate significant gains from the three-shot setting, indicating that few-shot learning is crucial for case manipulation. 

% The results from these intra-token probing tasks provide several key findings: (1) Larger models, such as llama3-70b and GPT-4 turbo, consistently outperform smaller models across all tasks. This indicates that increased model size contributes positively to handling complex token structure tasks.
% (2) Many tasks show significant performance improvements with few-shot learning, especially for smaller models. 

\begin{figure}[!ht] 
\centering
% \vspace{-1mm}
\subfigure[EM score performance on $k$-shot evaluation.] { \label{fig:inter-em}
\includegraphics[width=\linewidth]{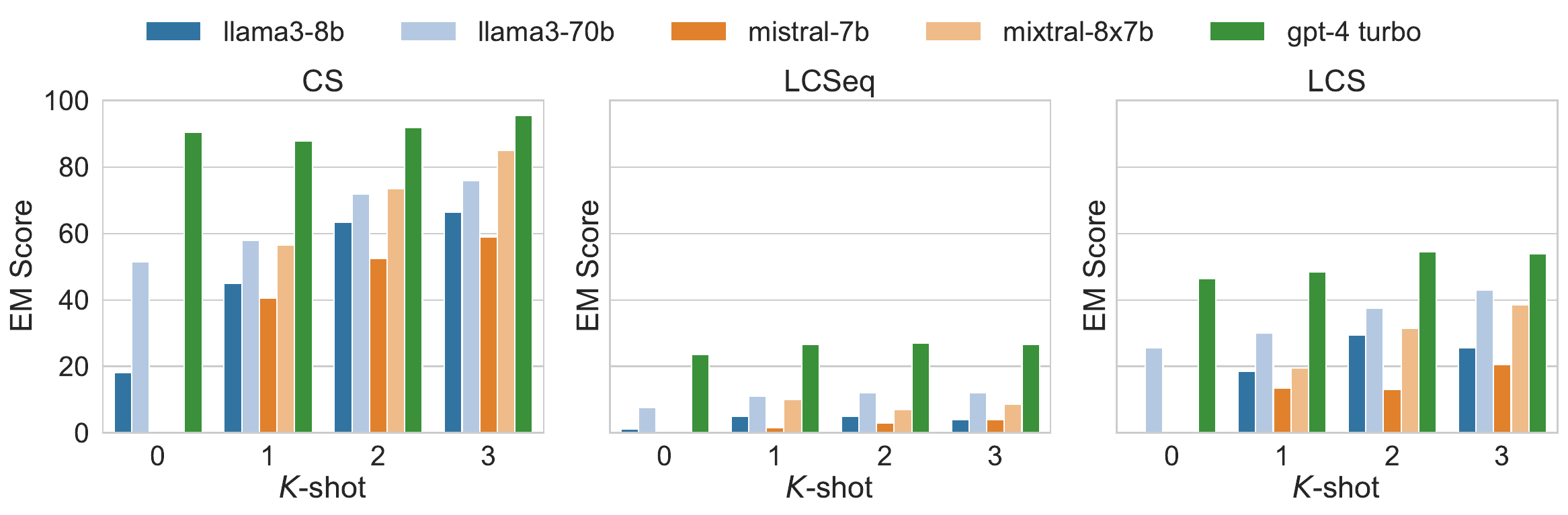}
}
\subfigure[Edit distance on $k$-shot evaluation.] { \label{fig:inter-ed}
\includegraphics[width=\linewidth]{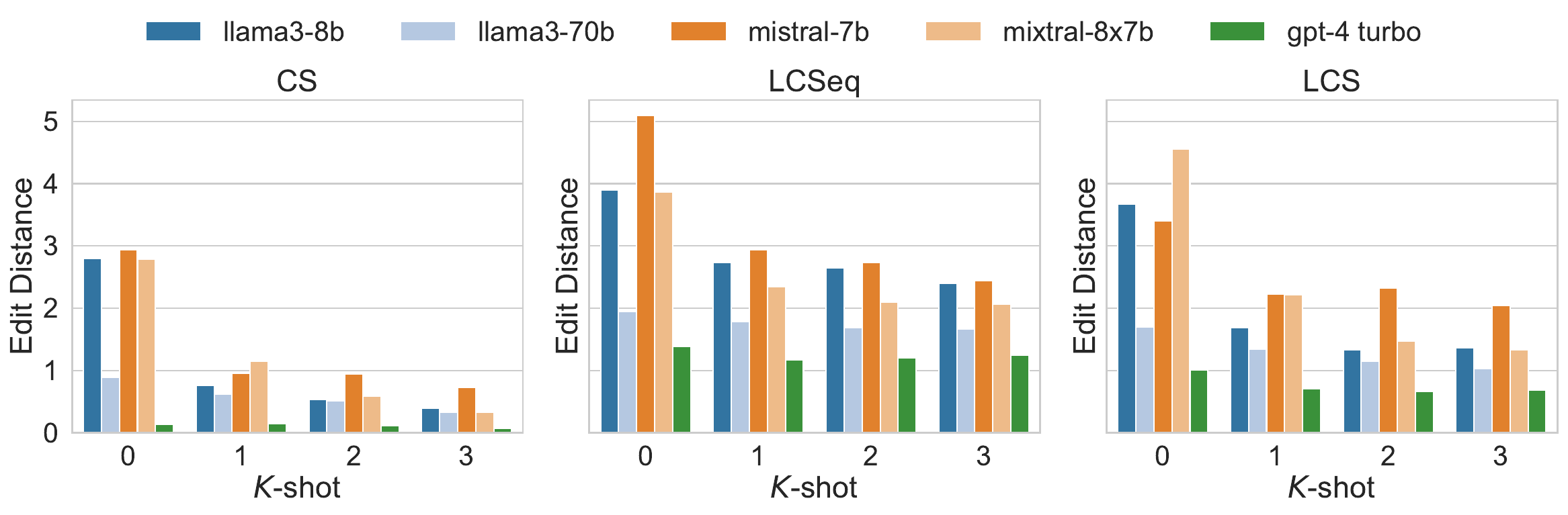}
}
% \vspace{-2mm}
\caption{$K$-shot performance on various inter-token probing tasks. For edit distances, lower is better.}
\label{fig:inter-token-probe}
% \vspace{-4mm}
\end{figure}

\subsection{Inter-Token Probing}
\label{sec:inter-token}

\paragraph{Task Description and Settings} To evaluate the capabilities of LLMs in understanding and manipulating relationships between multiple tokens, we devised inter-token probing tasks. These tasks focus on identifying common patterns and sequences across tokens, and they assess the model's ability to recognize and process such relationships. Specifically, we include three tasks:

\begin{itemize}[noitemsep, left=0pt, labelsep=4pt, topsep=0pt, partopsep=0pt]
\item \textbf{Common Substrings (CS)} -- The model identifies multiple common substrings between two given words.
\item \textbf{Longest Common Substrings (LCS)} -- The model identifies the longest continuous common substring for two given words.
\item \textbf{Longest Common Subsequences (LCSeq)} -- The model identifies the longest common subsequence (not necessarily continuous) between two given words.
\end{itemize}

For each task, we conduct many-shot evaluations (0-3 shot) and report the EM and edit distance (ED) score to test the model's ability to understand and manipulate relationships between tokens at a higher level. For CS tasks, the model's response is considered correct if it generates one of the multiple possible common substrings. Detailed test examples are provided in Appendix~\ref{ap:example-inter-token}.

% \comm{add results table or barplot here}

\paragraph{Results and Analysis}The results are presented in Figure~\ref{fig:inter-token-probe}.
The results for CS tasks indicate that larger models, such as Llama3-70B and GPT-4 Turbo, perform significantly better than smaller models. GPT-4 Turbo achieves the highest accuracy across all shot settings, demonstrating the model's robustness in identifying continuous substrings. Notably, Llama3-70B also shows strong performance, particularly in the three-shot setting. Sparse models like Mixtral-8x7B exhibit notable improvements compared to dense models, highlighting the effectiveness of sparse architectures in handling complex token relationships.

For LCS tasks, GPT-4 Turbo leads in performance, achieving high accuracy across all shot settings. Llama3-70B and Mixtral-8x7B show considerable improvements with increased shots, indicating that exposure to more examples helps models better identify multiple common substrings. Dense models like Mistral-7B lag behind, reinforcing the advantage of sparse architectures in such tasks.

The LCSeq task reveals that even the best-performing models face challenges with non-continuous patterns. While Llama3-70B and GPT-4 Turbo demonstrate superior performance, there is a significant drop in accuracy compared to CS tasks. Few-shot learning significantly enhances the performance of smaller models, such as Llama3-8B, which improves from 1\% in zero-shot to 4\% in three-shot settings. This underscores the importance of few-shot examples in aiding models to recognize and process non-continuous patterns.

% \paragraph{Discussion} 

\section{Typographical Variation}
\label{sec:rq3}

\begin{figure*}[!ht] 
\centering
% \vspace{-2mm}
\subfigure[TruthfulQA] { \label{fig:truthfulqa}
\includegraphics[width=0.48\linewidth]{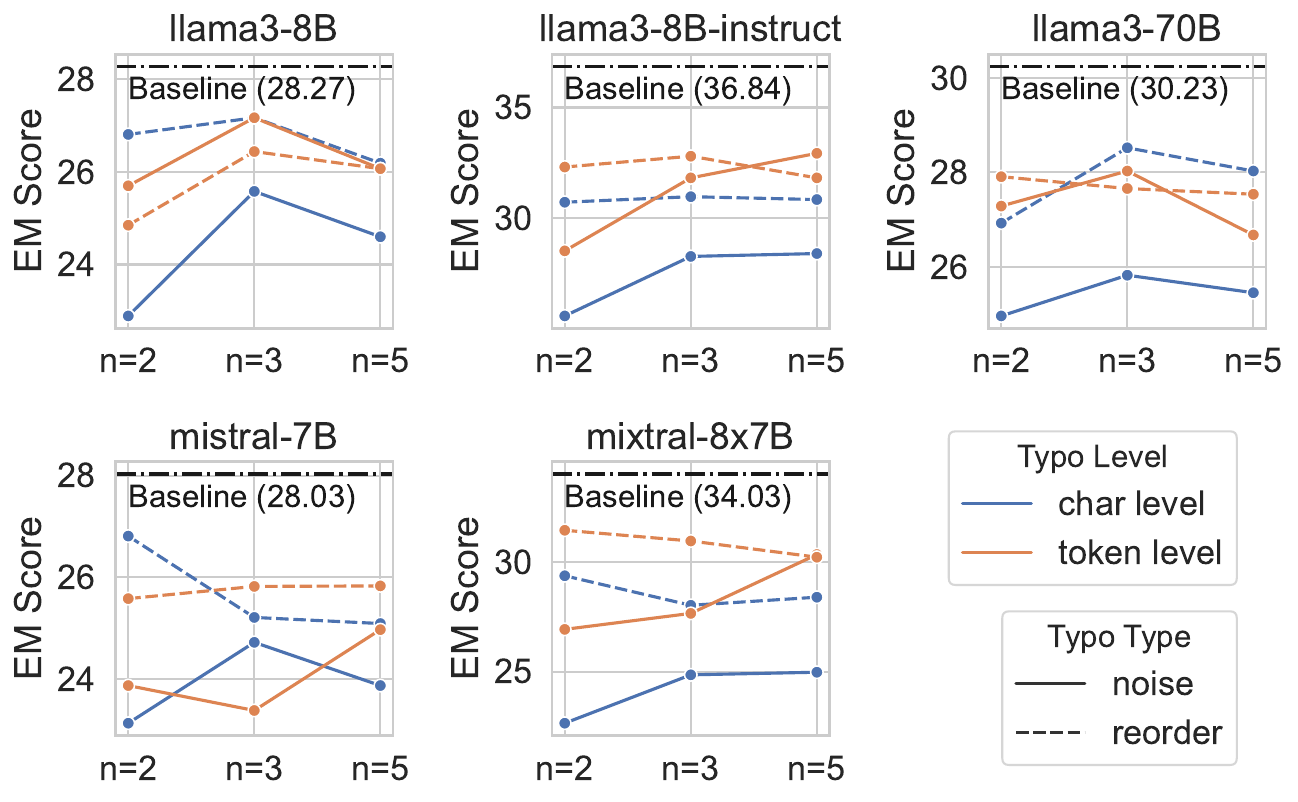}
}
% \vspace{-2mm}
\hfill
\subfigure[MMLU] { \label{fig:mmlu}
\includegraphics[width=0.48\linewidth]{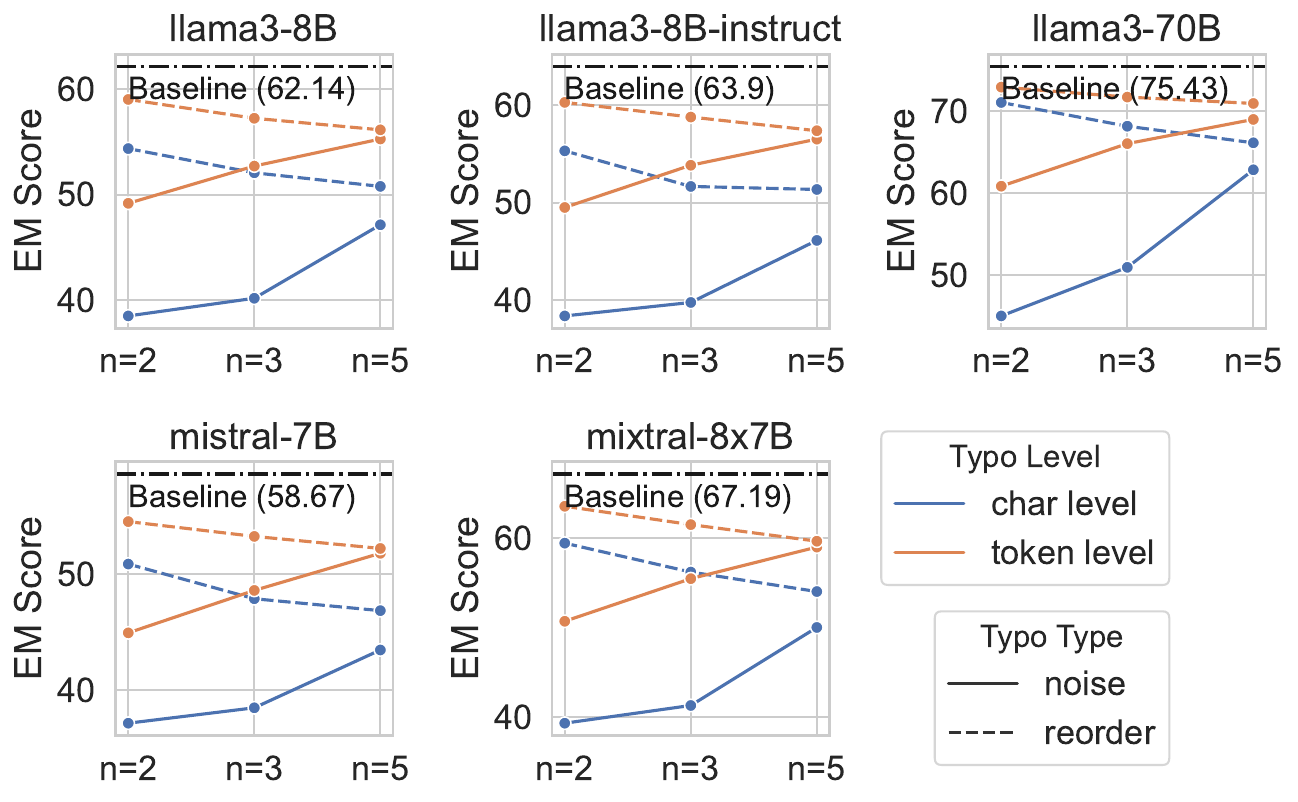}
}
\hfill 
% \vspace{-1mm}
\subfigure[GSM8K (5-shot)] { \label{fig:gsm8k}
\includegraphics[width=0.48\linewidth]{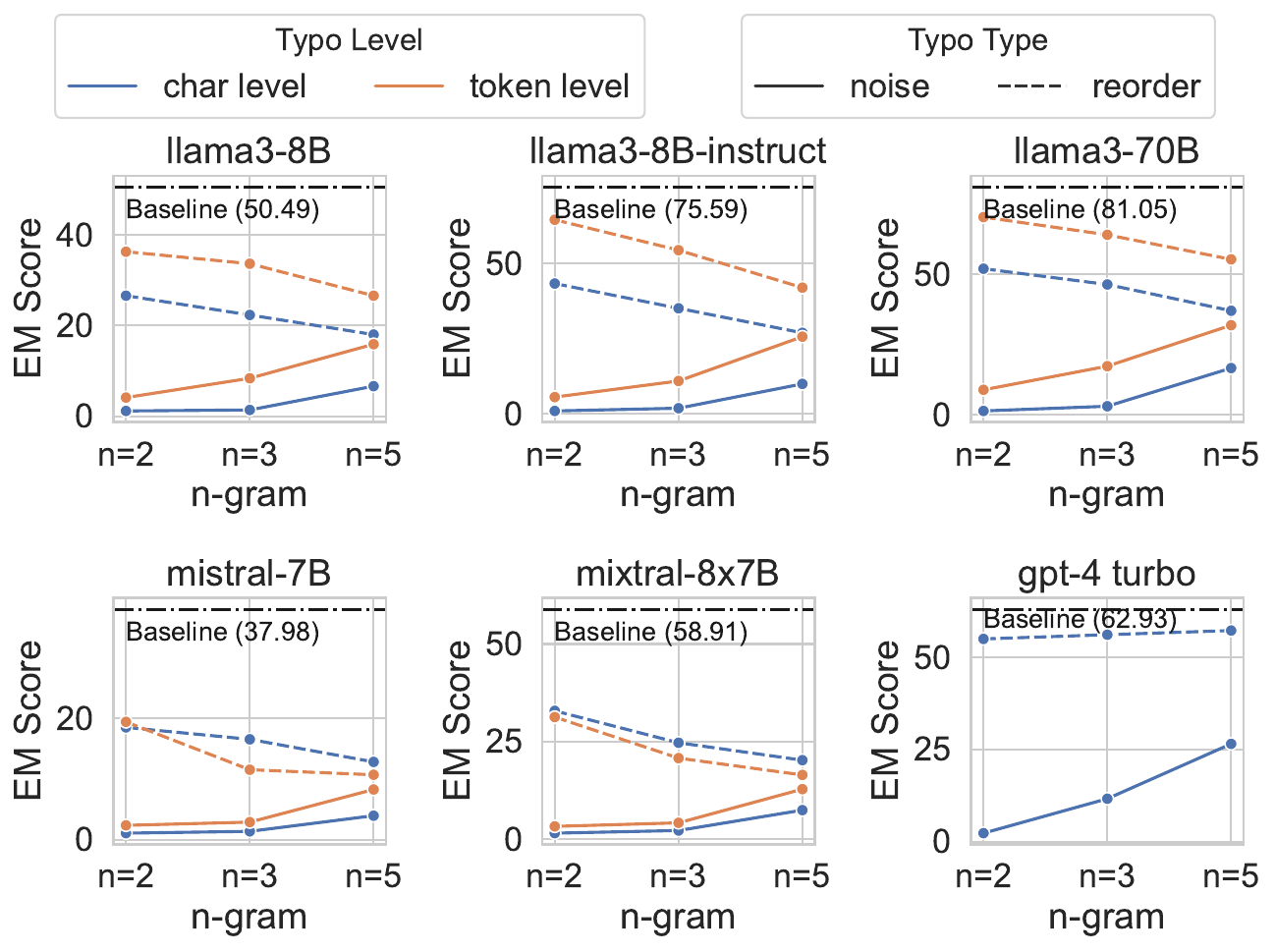}
}
\hfill
\subfigure[HumanEval] { \label{fig:humaneval}
\includegraphics[width=0.48\linewidth]{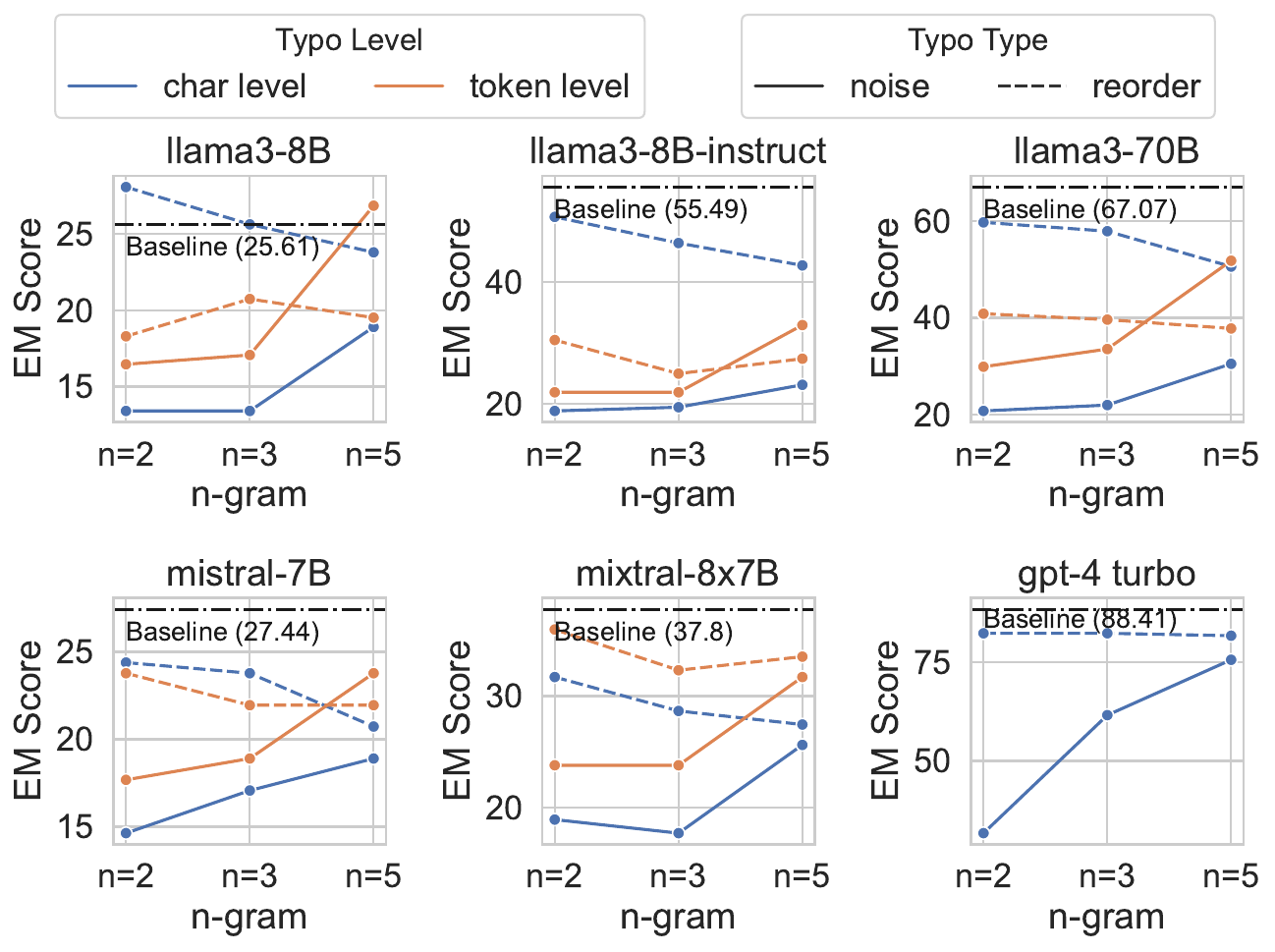}
}
% \vspace{-2mm}
\caption{Performance comparison for various models across different $n$-gram sizes ($n$=2,3,5) and typographical variations on (1) \textbf{TruthfulQA}, (b) \textbf{MMLU}, (c) \textbf{GSM8K}, and (d) \textbf{HumanEval}. The typographical variations include character-level (blue) and token-level (orange) perturbations, with noise (solid line) and reorder (dashed line) types. Baseline performance is indicated with a dotted line at the top of each plot.}
\label{fig:exp3}
% \vspace{-15pt}
\end{figure*}

To evaluate the robustness of LLMs to typographical variations, we constructed tasks that introduce character-level and token-level typographical errors into the input text. These tasks are designed to test the models' ability to maintain semantic understanding despite the presence of such errors. The datasets include MMLU~\cite{mmlu}, TruthfulQA~\cite{truthfulqa}, GSM8K~\cite{gsm8k}, and HumanEval~\cite{he}, ensuring diverse coverage.

\paragraph{Task Description and Settings} 
The primary goal of these tasks is to assess the LLMs' resilience to typographical errors at both the character and token levels, examining whether these models can maintain semantic understanding when faced with such perturbations. 
For \textbf{character-level typographical variation}, we employed $n$-gram shuffling within word boundaries (with $n$ set to 2, 3, 5) with a 50\% probability, and $n$-gram noise, which involve adding, deleting, and replacing characters, spaces, and punctuation marks to simulate spelling noise. This corruption occurs with a 30\% probability, including insertion, deletion, or substitution operations. \textbf{Token-level typographical variation} was introduced by shuffling tokens within $n$-grams of sizes 2, 3, and 5, with a 50\% probability of permutation or typo generation, similar to the character-level method. We include four tasks: 
\begin{itemize}[noitemsep, left=0pt, labelsep=4pt, topsep=0pt, partopsep=0pt]
\item \textbf{Character-Level Permutation}: Shuffling characters within word boundaries using $n$-grams of sizes 2, 3, and 5, with a 50\% probability.
\item \textbf{Character-Level Noise}: Adding, deleting, replacing random characters to simulate spelling noise, each with a 10\% probability.
\item \textbf{Token-Level Permutation}: Randomly reordering tokens using $n$-grams of sizes 2, 3, 5, with a 50\% probability.
\item \textbf{Token-Level Noise}: Adding, deleting, replacing tokens to inject noise, with a 30\% probability.
\end{itemize}

% Specifically, the permutation probability was set at 50\%. Noise addition occurred with a 30\% probability, split into 10\% for adding, deleting, and replacing characters. For additions, there was a 50\% chance the added character was randomly selected from the current n-gram and a 50\% chance it was selected from the full vocabulary.

For evaluation, we report pass@1 for HumanEval using a temperature of 0.2 and a top-$p$ of 0.95. For others, we used a temperature of 0. GSM8K was evaluated using a 5-shot setting, while MMLU, TruthfulQA, and HumanEval were assessed in a zero-shot setting. We measured performance on MMLU and TruthfulQA using perplexity for multiple-choice selection\footnote{Since GPT-4 Turbo does not support perplexity computation, it was excluded from the evaluation for these two tasks.}.

\paragraph{Results and Analysis} Figures \ref{fig:exp3} presents
% ~\ref{fig:truthfulqa}, \ref{fig:mmlu}, \ref{fig:gsm8k}, and \ref{fig:humaneval} present 
a comprehensive analysis of the impact of typographical variations on various LLMs, specifically focusing on character-level and token-level perturbations across different $n$-gram sizes ($n$=2, 3, 5). The evaluation covers a range of datasets, including TruthfulQA, MMLU, GSM8K (5-shot), and HumanEval.

Across all datasets and models, there is a consistent trend showing that LLMs are much more sensitive to noise (solid lines) than to reordering (dashed lines). Noise injection, which involves adding, deleting, or replacing characters or tokens, leads to more pronounced variations and generally degrades overall performance. This is evident from the lower EM scores for noise perturbations compared to reorder perturbations.

Despite the challenges posed by $n$-gram reordering within word boundaries, GPT-4 Turbo maintained high accuracy across all $n$-gram sizes. Character-level $n$-gram noise injection, simulating realistic spelling noise, further tested the models' robustness. The results indicate that all models experienced evident performance degradation, regardless of the parameter sizes, highlighting their sensitivity to typographical noise.

 For most models and datasets, as the $n$-gram size of noise injection increases from $n$=2 to 5, the performance tends to stabilize or improve. This trend suggests that models can better handle larger $n$-gram noise injection, likely because the context within larger $n$-grams provides more semantic coherence compared to smaller $n$-grams.

At the token level, models were subjected to $n$-gram permutations similar to those applied at the character level. The results indicated that models generally performed better with token-level permutations than with character-level shuffles and noise injection. This suggests that token-level errors may be less disruptive to the overall semantic structure of the input text.
% , allowing models to preserve meaning more effectively. 
% The ability to handle token-level perturbations more efficiently highlights the importance of maintaining the integrity of tokens within the context of the input text.

\section{Does BPE-dropout Matter?}
\label{sec:ft}
% \subsection{}

\begin{figure}[!ht]
    \vspace{-1mm}
    \centering
    \includegraphics[width=0.95\linewidth]{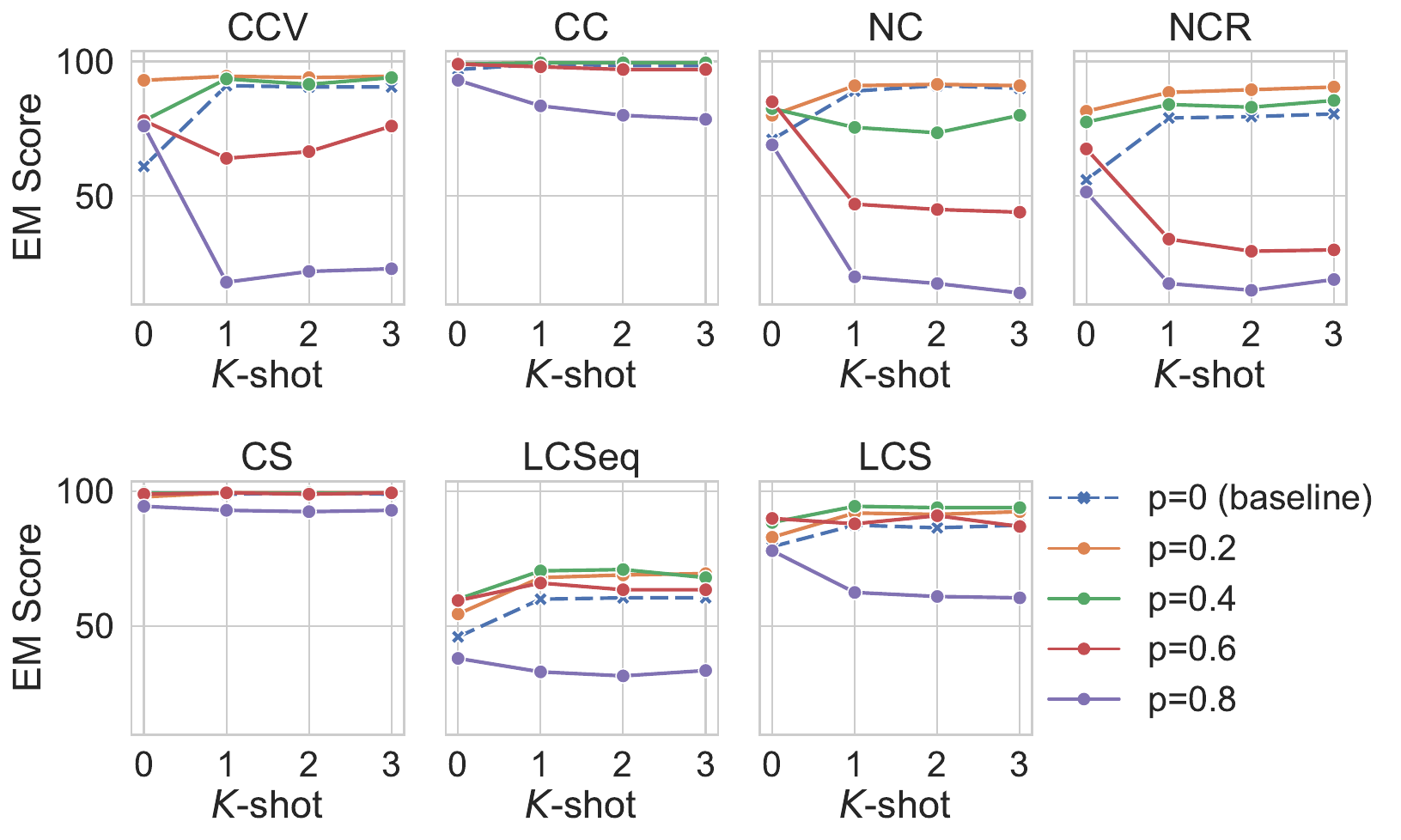}
     % \vspace{-0.5em}
    \caption{$K$-shot performance on various tasks (CCV, CC, NC, NCR, CS, LCSeq, and LCS) using the Mistral-7B model fine-tuned with a BPE-dropout tokenizer at different dropout rates $p$, ranging from 0 to 0.8. The baseline without BPE-dropout (\emph{i.e.}, $p=0$) is depicted with a dashed line. 
    It demonstrates that introducing a moderate amount of variability during tokenization improves the model's generalization capabilities, mitigating \textit{the curse of tokenization} issues.
    }
    \label{fig:tk-ft} 
    % \vspace{-8pt}
\end{figure}

\begin{figure*}[!ht] 
\centering
% \vspace{-1mm}
\subfigure[0-shot] { \label{fig:epoch_0_shot}
\includegraphics[width=0.48\linewidth]{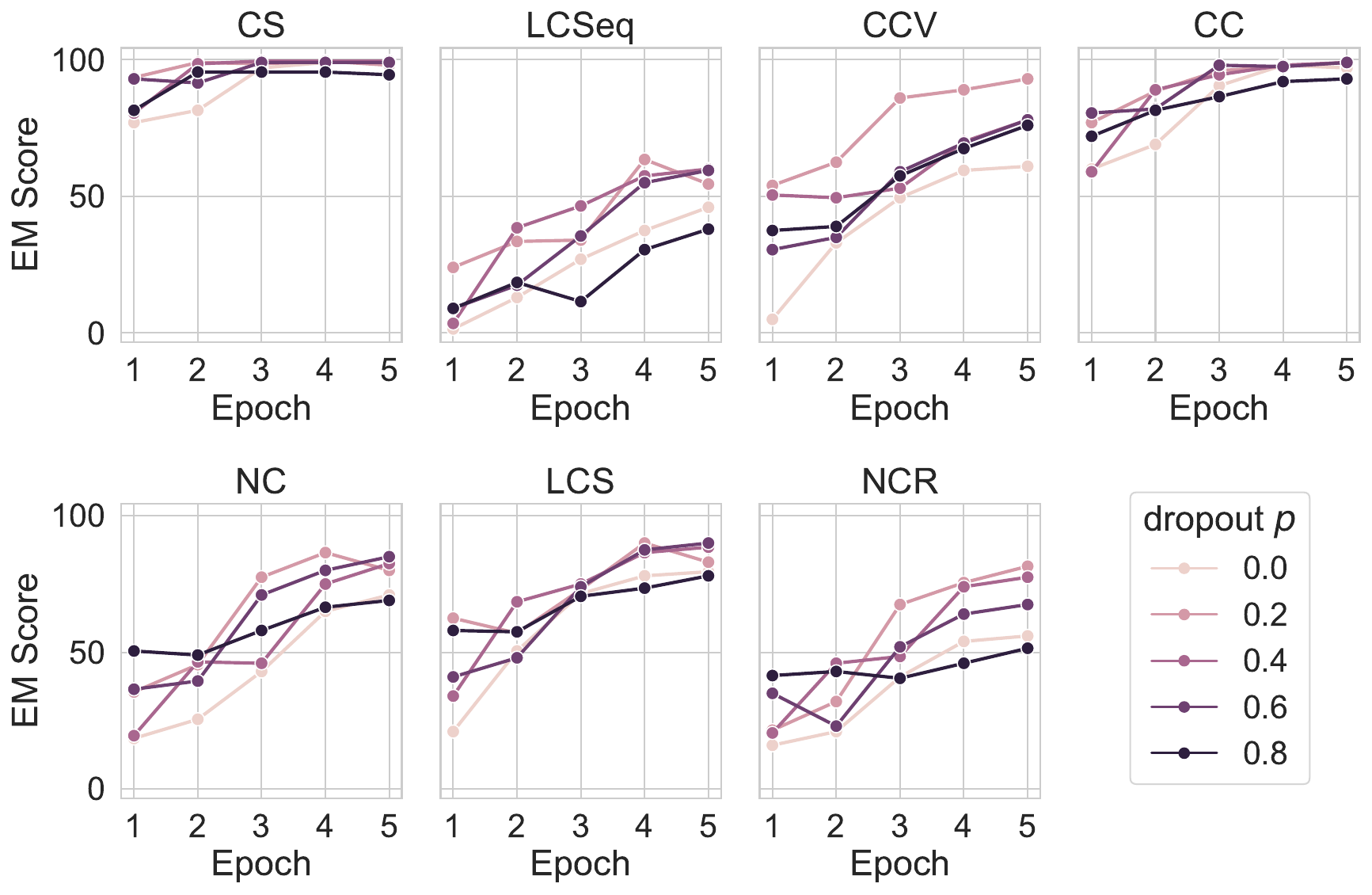}
}
% \vspace{-1mm}
\hfill
\subfigure[1-shot] { \label{fig:epoch_1_shot}
\includegraphics[width=0.48\linewidth]{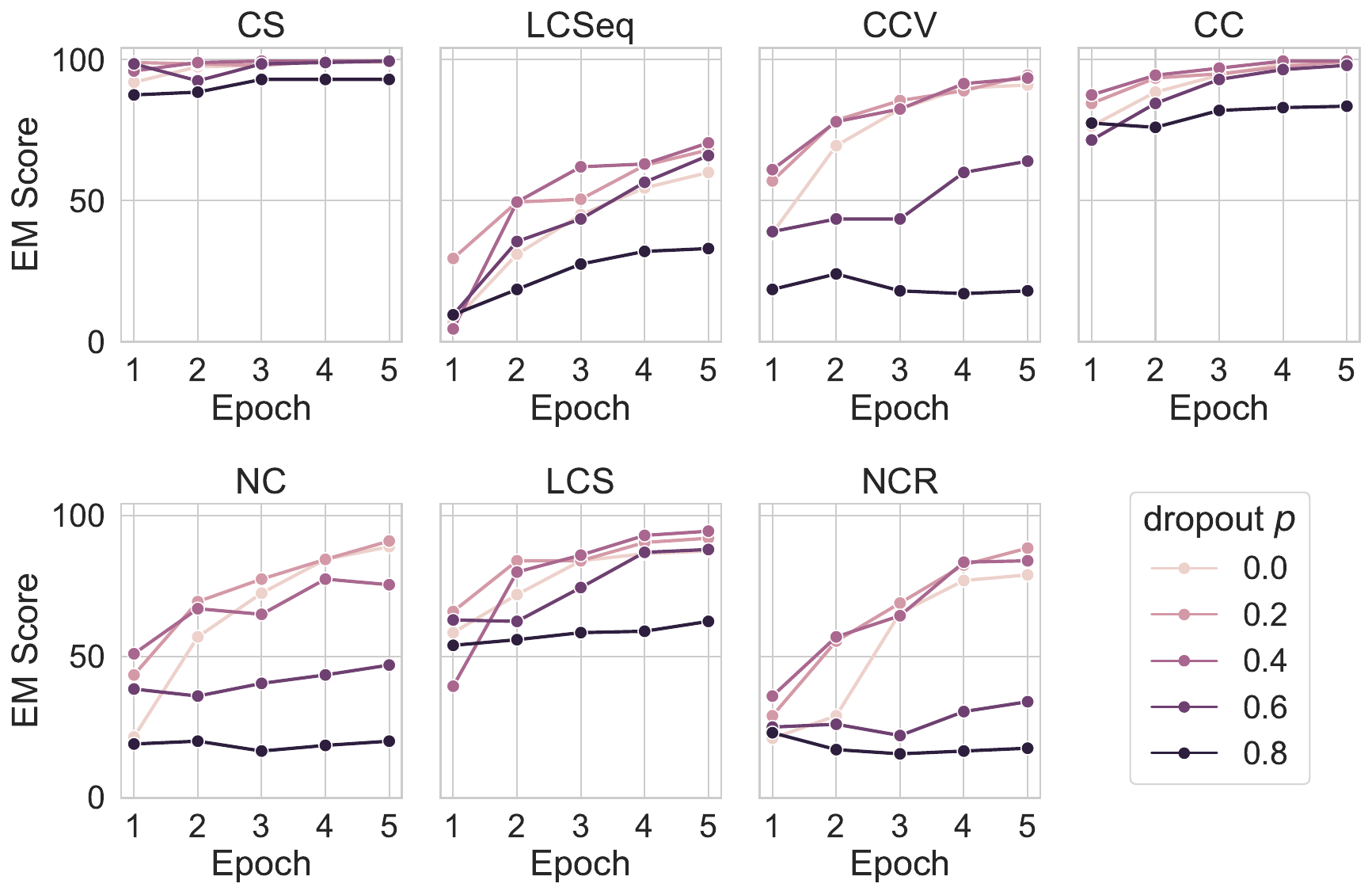}
}
\subfigure[2-shot] { \label{fig:2-shot}
\includegraphics[width=0.48\linewidth]{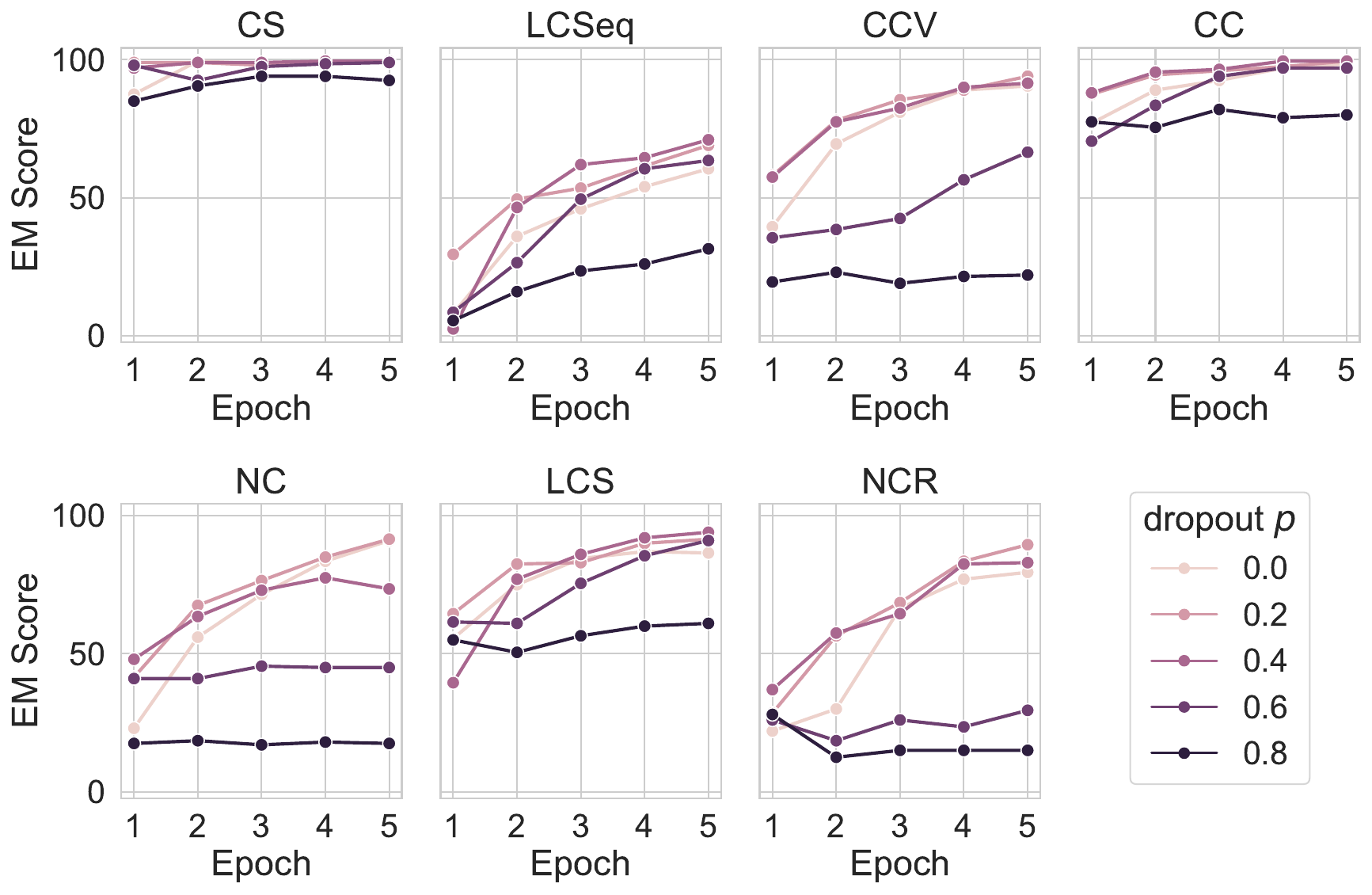}
}
\subfigure[3-shot] { \label{fig:3-shot}
\includegraphics[width=0.48\linewidth]{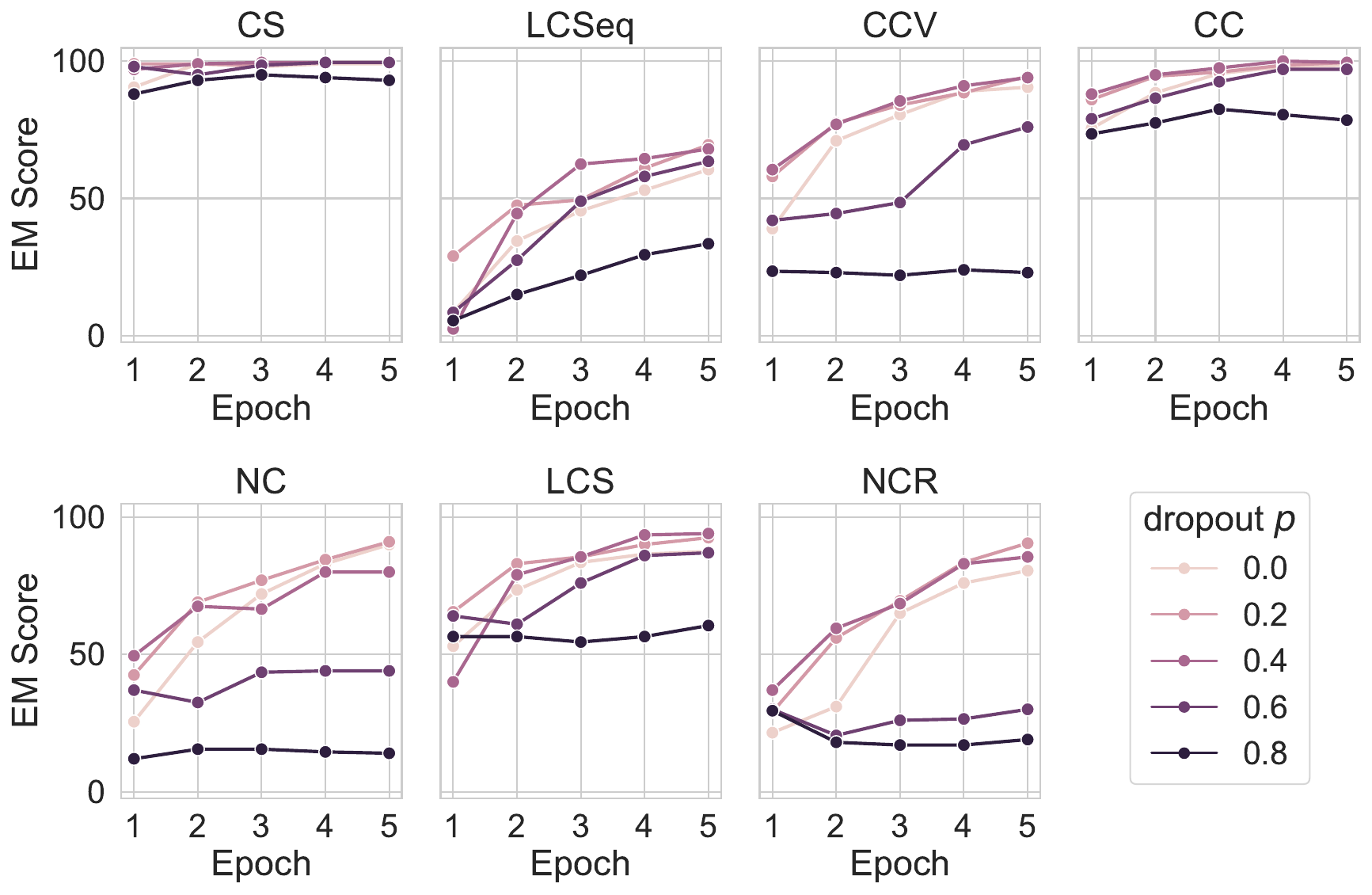}
}
\vspace{-1mm}
\caption{The impact of BPE-dropout on EM scores across seven tasks (CS, LCSeq, CCV, CC, NC, LCS, NCR) under different post-training conditions: (a) 0-shot, (b) 1-shot, (c) 2-shot, and (d) 3-shot. The dropout rates range from 0.0 to 0.8. The plots show that moderate dropout rates generally lead to improvements. Tasks such as CS and CC are more robust to dropout, maintaining higher scores even at moderate dropout rates, while tasks like NC, LCS, and NCR show significant performance drops with increasing dropout.}
\label{fig:ft_epoch}
% \vspace{-1mm}
\end{figure*}

To further enhance the robustness of LLMs, we explore regularized tokenization approach, BPE-dropout~\cite{provilkov-etal-2020-bpe}, which randomly drops BPE merges during tokenization. This technique allows text sequences to be tokenized in more diverse ways, promoting robustness to various token combinations and increasing the likelihood of encountering smaller tokens. Intuitively, this diversity benefits the model's understanding of internal token structures.

\paragraph{Training Setup}
For the training data, we synthesized a dataset consisting of 111k examples specifically designed for RQ2. We employ the AdamW optimizer~\cite{loshchilov2017decoupled} with the hyperpameters of $\beta_1=0.9$, $\beta_2=0.95$. The peak learning rate is set to 5e-5, and the minimum learning rate is set to 1e-6. The learning rate warms up during the first 10\% of training steps and then decays with a cosine scheduler.

Given the difference in data distribution resulting from BPE-dropout, we post-train Mistral-7B~\cite{mistral23} on the training split for 5 epochs with a global batch size of 16. Following the pre-training recipe, we concatenate all sequences and then chunk them into fixed context lengths of 4096 for our autoregressive post-training. We conduct data shuffling within the same epochs. More detailed training settings are provided in Appendix~\ref{ap:train_details}.

% The training process was conducted over 5 epochs, with a peak learning rate set at 5e-5 and a minimum learning rate of 1e-6. We employed a global batch size of 16. 

% \comm{add final results}
\paragraph{Results and Analysis}
Figure~\ref{fig:tk-ft} shows the impact of varying BPE-dropout rates on the Mistral-7B model's performance across multiple $K$-shot settings and tasks.  
The baseline performance, with a dropout rate of $p=0$, shows robust results across several tasks, particularly in CS and CC. These tasks are relatively straightforward, involving simple token manipulations that do not significantly challenge the model’s capacity to generalize from zero-shot to few-shot scenarios. The high performance in these tasks suggests that the Mistral-7B model, even without BPE-dropout, is adept at handling simpler token relationships. However, it is important to note that the baseline performance does not uniformly extend to more complex tasks.

In contrast, tasks such as LCSeq reveal relatively low performance across all models, irrespective of the BPE-dropout rate. This suggests inherent difficulties in these tasks that stem from the requirement to identify non-continuous and intricate token patterns. The consistent under-performance indicates that LCSeq tasks pose a significant challenge to the model’s ability to generalize, likely due to the increased complexity in recognizing and processing longer and fragmented sequences.

Interestingly, the introduction of a moderate BPE-dropout rate ($p=0.2$) frequently surpasses the baseline, highlighting the benefits of inducing variability during tokenization. This moderate dropout rate enhances the model’s generalization capabilities by preventing overfitting and promoting a more robust learning process. Notably, in tasks such as CCV, NC, and LCS, the $p=0.2$ model consistently achieves higher EM scores, underscoring the benefits of incorporating tokenization regularization.

Our analysis reveals that higher dropout rates ($p=0.6$ and $p=0.8$) exhibit relatively lower performance across most tasks. This decline can be attributed to insufficient training, as the dataset was trained for only five epochs. The higher dropout rates introduce greater tokenization variation, which necessitates additional training compute to achieve convergence. The lack of adequate training epochs likely hindered these models from fully leveraging the potential benefits of higher BPE-dropout rates. Moreover, we report the test-set performance across seven tasks over the course of BPE-dropout fine-tuning in Figure~\ref{fig:ft_epoch}. We provided detailed training analysis in Appendix \S\ref{ap:results}.

% Furthermore, the performance degradation observed with higher dropout rates highlights the delicate balance required in applying BPE-dropout. While moderate dropout rates can enhance generalization, excessive dropout may lead to too much variability, thereby complicating the learning process and requiring more extensive computational resources to converge effectively. This finding emphasizes the importance of fine-tuning the dropout rate to optimize model performance without overburdening the training process. 

% \comm{add epoch vs. perf. line plot (trend)}

\section{Conclusion}
\label{sec:concl}
In this study, we investigated the challenge of \textit{the curse of tokenization}, comprehensively evaluating mainstream LLMs across thirteen tasks sensitive to conventional subword tokenization. Our findings reveal that while larger models and increased shot counts can partially mitigate these issues, LLMs still struggle with understanding internal structures and token compositions. Moderate BPE-dropout can alleviate some of these challenges, whereas larger drop rates lead to performance degradation. We encourage the research community to develop more flexible approaches to further address these limitations and enhance model robustness.

\section*{Limitations}
While our study provides valuable insights into the robustness and performance of large language models (LLMs) under various tokenization and typographical variation scenarios, several limitations should be acknowledged:

\paragraph{Data Diversity and Size} The training data synthesized for RQ2 (token structure probe) consists of approximate 30k examples. While this dataset size is substantial, it may not fully capture the diversity and complexity of real-world text. Future work could benefit from expanding the dataset size and incorporating a wider range of linguistic phenomena.

\paragraph{Evaluation Metrics} Our evaluation primarily relies on metrics such as pass@1 for HumanEval and accuracy for MMLU, TruthfulQA, and GSM8K. While these metrics provide valuable insights, they may not fully capture the nuanced performance of LLMs in real-world applications. Incorporating additional metrics that assess other aspects of model performance, such as robustness to out-of-distribution data and interpretability, could provide a more comprehensive evaluation.

\paragraph{Subword Regularization} Although our use of BPE-dropout shows promising improvements in model robustness and accuracy, the approach introduces randomness into the tokenization process. This randomness can lead to variability in model performance, making it challenging to ensure consistent improvements across different datasets and tasks. Further research is needed to optimize the BPE-dropout technique and evaluate its long-term impact on model performance.

\paragraph{Typographical Variation} Our study focuses on character-level and token-level typographical variations, but it does not address other common types of text perturbations, such as grammatical errors, semantic variations, or contextual inconsistencies. Exploring the effects of these additional types of variations could provide a more holistic understanding of LLM robustness.

\paragraph{Generalizability} The findings from our evaluation on specific datasets (MMLU, TruthfulQA, GSM8K, and HumanEval) may not generalize to all types of text and tasks. Further studies are needed to assess the generalizability of our findings across a broader range of datasets and real-world scenarios, such as extending to multilingual evaluation~\cite{peng-etal-2024-humaneval-xl}.

% \paragraph{Model Scalability} While larger models like GPT-4 Turbo and Llama3-70b demonstrate superior performance, their scalability and resource requirements pose significant challenges. Training and deploying these large models require substantial computational resources, which may not be feasible for all researchers and practitioners. Investigating ways to achieve similar levels of performance with smaller, more efficient models could help mitigate these challenges.

% By acknowledging these limitations, we aim to provide a balanced perspective on our findings and highlight areas for future research. 

\section*{Ethical Consideration}
\paragraph{Bias and Fairness} Tokenization strategies can introduce or exacerbate biases present in the training data. Our study, which involves diverse tokenization techniques like BPE-Dropout, should include thorough bias assessments to ensure that these methods do not perpetuate unfair or discriminatory outcomes. Mitigating bias is crucial for creating fair and equitable AI systems.

\paragraph{Transparency and Interpretability} Our research involves complex tokenization processes that can obscure the decision-making of LLMs. Enhancing the transparency of these models by providing clear explanations of how tokenization impacts model behavior is essential. This transparency helps build trust and allows users to understand and identify potential issues in language model predictions.

\paragraph{Privacy and Data Security} The datasets used for training and evaluating tokenization methods often contain sensitive information. Ensuring data anonymization and compliance with data protection regulations is critical to protecting user privacy. Our study adheres to strict data security protocols to prevent any misuse of sensitive information.

\section*{Acknowledgements}
We would like to thank all anonymous reviewers for their insightful comments and feedback. Qiwei Peng is supported by DisAI - Improving scientific excellence and creativity in combating disinformation with artificial intelligence and language technologies, a project funded by European Union under the Horizon Europe, GA No. 101079164.

\bibliography{section/custom}

\begin{thebibliography}{37}
\expandafter\ifx\csname natexlab\endcsname\relax\def\natexlab#1{#1}\fi

\bibitem[{Abdou et~al.(2022)Abdou, Ravishankar, Kulmizev, and S{\o}gaard}]{abdou-etal-2022-word}
Mostafa Abdou, Vinit Ravishankar, Artur Kulmizev, and Anders S{\o}gaard. 2022.
\newblock \href {https://doi.org/10.18653/v1/2022.acl-long.476} {Word order does matter and shuffled language models know it}.
\newblock In \emph{Proceedings of the 60th Annual Meeting of the Association for Computational Linguistics (Volume 1: Long Papers)}, pages 6907--6919, Dublin, Ireland. Association for Computational Linguistics.

\bibitem[{Aghajanyan et~al.(2022)Aghajanyan, Okhonko, Lewis, Joshi, Xu, Ghosh, and Zettlemoyer}]{htlm22}
Armen Aghajanyan, Dmytro Okhonko, Mike Lewis, Mandar Joshi, Hu~Xu, Gargi Ghosh, and Luke Zettlemoyer. 2022.
\newblock \href {https://openreview.net/forum?id=P-pPW1nxf1r} {{HTLM:} hyper-text pre-training and prompting of language models}.
\newblock In \emph{The Tenth International Conference on Learning Representations, {ICLR} 2022, Virtual Event, April 25-29, 2022}. OpenReview.net.

\bibitem[{AI@Meta(2024)}]{llama3modelcard}
AI@Meta. 2024.
\newblock \href {https://github.com/meta-llama/llama3/blob/main/MODEL_CARD.md} {Llama 3 model card}.

\bibitem[{Anil et~al.(2023)Anil, Borgeaud, Wu, Alayrac, Yu, Soricut, Schalkwyk, Dai, Hauth, Millican, Silver, Petrov, Johnson, Antonoglou, Schrittwieser, Glaese, Chen, Pitler, Lillicrap, Lazaridou, Firat, Molloy, Isard, Barham, Hennigan, Lee, Viola, Reynolds, Xu, Doherty, Collins, Meyer, Rutherford, Moreira, Ayoub, Goel, Tucker, Piqueras, Krikun, Barr, Savinov, Danihelka, Roelofs, White, Andreassen, von Glehn, Yagati, Kazemi, Gonzalez, Khalman, Sygnowski, and et~al.}]{gemini23}
Rohan Anil, Sebastian Borgeaud, Yonghui Wu, Jean{-}Baptiste Alayrac, Jiahui Yu, Radu Soricut, Johan Schalkwyk, Andrew~M. Dai, Anja Hauth, Katie Millican, David Silver, Slav Petrov, Melvin Johnson, Ioannis Antonoglou, Julian Schrittwieser, Amelia Glaese, Jilin Chen, Emily Pitler, Timothy~P. Lillicrap, Angeliki Lazaridou, Orhan Firat, James Molloy, Michael Isard, Paul~Ronald Barham, Tom Hennigan, Benjamin Lee, Fabio Viola, Malcolm Reynolds, Yuanzhong Xu, Ryan Doherty, Eli Collins, Clemens Meyer, Eliza Rutherford, Erica Moreira, Kareem Ayoub, Megha Goel, George Tucker, Enrique Piqueras, Maxim Krikun, Iain Barr, Nikolay Savinov, Ivo Danihelka, Becca Roelofs, Ana{\"{\i}}s White, Anders Andreassen, Tamara von Glehn, Lakshman Yagati, Mehran Kazemi, Lucas Gonzalez, Misha Khalman, Jakub Sygnowski, and et~al. 2023.
\newblock \href {https://doi.org/10.48550/ARXIV.2312.11805} {Gemini: {A} family of highly capable multimodal models}.
\newblock \emph{CoRR}, abs/2312.11805.

\bibitem[{Ben~Allal et~al.(2022)Ben~Allal, Muennighoff, Kumar~Umapathi, Lipkin, and von Werra}]{bigcode-evaluation-harness}
Loubna Ben~Allal, Niklas Muennighoff, Logesh Kumar~Umapathi, Ben Lipkin, and Leandro von Werra. 2022.
\newblock A framework for the evaluation of code generation models.
\newblock \url{https://github.com/bigcode-project/bigcode-evaluation-harness}.

\bibitem[{Brown et~al.(2020)Brown, Mann, Ryder, Subbiah, Kaplan, Dhariwal, Neelakantan, Shyam, Sastry, Askell et~al.}]{brown2020language}
Tom Brown, Benjamin Mann, Nick Ryder, Melanie Subbiah, Jared~D Kaplan, Prafulla Dhariwal, Arvind Neelakantan, Pranav Shyam, Girish Sastry, Amanda Askell, et~al. 2020.
\newblock Language models are few-shot learners.
\newblock \emph{Advances in neural information processing systems}, 33:1877--1901.

\bibitem[{Cao et~al.(2023)Cao, Kojima, Matsuo, and Iwasawa}]{CaoKMI23}
Qi~Cao, Takeshi Kojima, Yutaka Matsuo, and Yusuke Iwasawa. 2023.
\newblock \href {https://doi.org/10.18653/V1/2023.EMNLP-MAIN.550} {Unnatural error correction: {GPT-4} can almost perfectly handle unnatural scrambled text}.
\newblock In \emph{Proceedings of the 2023 Conference on Empirical Methods in Natural Language Processing, {EMNLP} 2023, Singapore, December 6-10, 2023}, pages 8898--8913. Association for Computational Linguistics.

\bibitem[{Chai(2021)}]{chai2021tokenization-PTMs}
Yekun Chai. 2021.
\newblock {Word Tokenization for Pre-trained Models}.
\newblock \url{https://cyk1337.github.io/notes/2021/11/29/Subword-Tokenization-in-NLP/}.

\bibitem[{Chai et~al.(2024)Chai, Liu, Xiao, Wang, Sun, and Wu}]{chai2024autoregressive}
Yekun Chai, Qingyi Liu, Jingwu Xiao, Shuohuan Wang, Yu~Sun, and Hua Wu. 2024.
\newblock Autoregressive pre-training on pixels and texts.
\newblock \emph{arXiv preprint arXiv:2404.10710}.

\bibitem[{Chai et~al.(2023)Chai, Wang, Pang, Sun, Tian, and Wu}]{ernie-code}
Yekun Chai, Shuohuan Wang, Chao Pang, Yu~Sun, Hao Tian, and Hua Wu. 2023.
\newblock \href {https://doi.org/10.18653/v1/2023.findings-acl.676} {{ERNIE}-code: Beyond {E}nglish-centric cross-lingual pretraining for programming languages}.
\newblock In \emph{Findings of the Association for Computational Linguistics: ACL 2023}, pages 10628--10650, Toronto, Canada. Association for Computational Linguistics.

\bibitem[{Chen et~al.(2021)Chen, Tworek, Jun, Yuan, de~Oliveira~Pinto, Kaplan, Edwards, Burda, Joseph, Brockman, Ray, Puri, Krueger, Petrov, Khlaaf, Sastry, Mishkin, Chan, Gray, Ryder, Pavlov, Power, Kaiser, Bavarian, Winter, Tillet, Such, Cummings, Plappert, Chantzis, Barnes, Herbert{-}Voss, Guss, Nichol, Paino, Tezak, Tang, Babuschkin, Balaji, Jain, Saunders, Hesse, Carr, Leike, Achiam, Misra, Morikawa, Radford, Knight, Brundage, Murati, Mayer, Welinder, McGrew, Amodei, McCandlish, Sutskever, and Zaremba}]{he}
Mark Chen, Jerry Tworek, Heewoo Jun, Qiming Yuan, Henrique~Pond{\'{e}} de~Oliveira~Pinto, Jared Kaplan, Harrison Edwards, Yuri Burda, Nicholas Joseph, Greg Brockman, Alex Ray, Raul Puri, Gretchen Krueger, Michael Petrov, Heidy Khlaaf, Girish Sastry, Pamela Mishkin, Brooke Chan, Scott Gray, Nick Ryder, Mikhail Pavlov, Alethea Power, Lukasz Kaiser, Mohammad Bavarian, Clemens Winter, Philippe Tillet, Felipe~Petroski Such, Dave Cummings, Matthias Plappert, Fotios Chantzis, Elizabeth Barnes, Ariel Herbert{-}Voss, William~Hebgen Guss, Alex Nichol, Alex Paino, Nikolas Tezak, Jie Tang, Igor Babuschkin, Suchir Balaji, Shantanu Jain, William Saunders, Christopher Hesse, Andrew~N. Carr, Jan Leike, Joshua Achiam, Vedant Misra, Evan Morikawa, Alec Radford, Matthew Knight, Miles Brundage, Mira Murati, Katie Mayer, Peter Welinder, Bob McGrew, Dario Amodei, Sam McCandlish, Ilya Sutskever, and Wojciech Zaremba. 2021.
\newblock \href {http://arxiv.org/abs/2107.03374} {Evaluating large language models trained on code}.
\newblock \emph{CoRR}, abs/2107.03374.

\bibitem[{Clark et~al.(2022)Clark, Garrette, Turc, and Wieting}]{clark-etal-2022-canine}
Jonathan~H. Clark, Dan Garrette, Iulia Turc, and John Wieting. 2022.
\newblock \href {https://doi.org/10.1162/tacl_a_00448} {Canine: Pre-training an efficient tokenization-free encoder for language representation}.
\newblock \emph{Transactions of the Association for Computational Linguistics}, 10:73--91.

\bibitem[{Cobbe et~al.(2021)Cobbe, Kosaraju, Bavarian, Chen, Jun, Kaiser, Plappert, Tworek, Hilton, Nakano, Hesse, and Schulman}]{gsm8k}
Karl Cobbe, Vineet Kosaraju, Mohammad Bavarian, Mark Chen, Heewoo Jun, Lukasz Kaiser, Matthias Plappert, Jerry Tworek, Jacob Hilton, Reiichiro Nakano, Christopher Hesse, and John Schulman. 2021.
\newblock \href {http://arxiv.org/abs/2110.14168} {Training verifiers to solve math word problems}.
\newblock \emph{CoRR}, abs/2110.14168.

\bibitem[{Gao et~al.(2023)Gao, Tow, Abbasi, Biderman, Black, DiPofi, Foster, Golding, Hsu, Le~Noac'h, Li, McDonell, Muennighoff, Ociepa, Phang, Reynolds, Schoelkopf, Skowron, Sutawika, Tang, Thite, Wang, Wang, and Zou}]{eval-harness}
Leo Gao, Jonathan Tow, Baber Abbasi, Stella Biderman, Sid Black, Anthony DiPofi, Charles Foster, Laurence Golding, Jeffrey Hsu, Alain Le~Noac'h, Haonan Li, Kyle McDonell, Niklas Muennighoff, Chris Ociepa, Jason Phang, Laria Reynolds, Hailey Schoelkopf, Aviya Skowron, Lintang Sutawika, Eric Tang, Anish Thite, Ben Wang, Kevin Wang, and Andy Zou. 2023.
\newblock \href {https://doi.org/10.5281/zenodo.10256836} {A framework for few-shot language model evaluation}.

\bibitem[{Hendrycks et~al.(2021)Hendrycks, Burns, Basart, Zou, Mazeika, Song, and Steinhardt}]{mmlu}
Dan Hendrycks, Collin Burns, Steven Basart, Andy Zou, Mantas Mazeika, Dawn Song, and Jacob Steinhardt. 2021.
\newblock \href {https://openreview.net/forum?id=d7KBjmI3GmQ} {Measuring massive multitask language understanding}.
\newblock In \emph{9th International Conference on Learning Representations, {ICLR} 2021, Virtual Event, Austria, May 3-7, 2021}. OpenReview.net.

\bibitem[{Jiang et~al.(2023)Jiang, Sablayrolles, Mensch, Bamford, Chaplot, de~Las~Casas, Bressand, Lengyel, Lample, Saulnier, Lavaud, Lachaux, Stock, Scao, Lavril, Wang, Lacroix, and Sayed}]{mistral23}
Albert~Q. Jiang, Alexandre Sablayrolles, Arthur Mensch, Chris Bamford, Devendra~Singh Chaplot, Diego de~Las~Casas, Florian Bressand, Gianna Lengyel, Guillaume Lample, Lucile Saulnier, L{\'{e}}lio~Renard Lavaud, Marie{-}Anne Lachaux, Pierre Stock, Teven~Le Scao, Thibaut Lavril, Thomas Wang, Timoth{\'{e}}e Lacroix, and William~El Sayed. 2023.
\newblock \href {https://doi.org/10.48550/ARXIV.2310.06825} {Mistral 7b}.
\newblock \emph{CoRR}, abs/2310.06825.

\bibitem[{Jiang et~al.(2024)Jiang, Sablayrolles, Roux, Mensch, Savary, Bamford, Chaplot, de~Las~Casas, Hanna, Bressand, Lengyel, Bour, Lample, Lavaud, Saulnier, Lachaux, Stock, Subramanian, Yang, Antoniak, Scao, Gervet, Lavril, Wang, Lacroix, and Sayed}]{mixtral24}
Albert~Q. Jiang, Alexandre Sablayrolles, Antoine Roux, Arthur Mensch, Blanche Savary, Chris Bamford, Devendra~Singh Chaplot, Diego de~Las~Casas, Emma~Bou Hanna, Florian Bressand, Gianna Lengyel, Guillaume Bour, Guillaume Lample, L{\'{e}}lio~Renard Lavaud, Lucile Saulnier, Marie{-}Anne Lachaux, Pierre Stock, Sandeep Subramanian, Sophia Yang, Szymon Antoniak, Teven~Le Scao, Th{\'{e}}ophile Gervet, Thibaut Lavril, Thomas Wang, Timoth{\'{e}}e Lacroix, and William~El Sayed. 2024.
\newblock \href {https://doi.org/10.48550/ARXIV.2401.04088} {Mixtral of experts}.
\newblock \emph{CoRR}, abs/2401.04088.

\bibitem[{Jozefowicz et~al.(2016)Jozefowicz, Vinyals, Schuster, Shazeer, and Wu}]{jozefowicz2016exploring}
Rafal Jozefowicz, Oriol Vinyals, Mike Schuster, Noam Shazeer, and Yonghui Wu. 2016.
\newblock Exploring the limits of language modeling.
\newblock \emph{arXiv preprint arXiv:1602.02410}.

\bibitem[{Kudo(2018)}]{kudo-2018-subword}
Taku Kudo. 2018.
\newblock \href {https://doi.org/10.18653/v1/P18-1007} {Subword regularization: Improving neural network translation models with multiple subword candidates}.
\newblock In \emph{Proceedings of the 56th Annual Meeting of the Association for Computational Linguistics (Volume 1: Long Papers)}, pages 66--75, Melbourne, Australia. Association for Computational Linguistics.

\bibitem[{Kudo and Richardson(2018)}]{spm18}
Taku Kudo and John Richardson. 2018.
\newblock \href {https://doi.org/10.18653/V1/D18-2012} {Sentencepiece: {A} simple and language independent subword tokenizer and detokenizer for neural text processing}.
\newblock In \emph{Proceedings of the 2018 Conference on Empirical Methods in Natural Language Processing, {EMNLP} 2018: System Demonstrations, Brussels, Belgium, October 31 - November 4, 2018}, pages 66--71. Association for Computational Linguistics.

\bibitem[{Lin et~al.(2022)Lin, Hilton, and Evans}]{truthfulqa}
Stephanie Lin, Jacob Hilton, and Owain Evans. 2022.
\newblock \href {https://doi.org/10.18653/v1/2022.acl-long.229} {{T}ruthful{QA}: Measuring how models mimic human falsehoods}.
\newblock In \emph{Proceedings of the 60th Annual Meeting of the Association for Computational Linguistics (Volume 1: Long Papers)}, pages 3214--3252, Dublin, Ireland. Association for Computational Linguistics.

\bibitem[{Loshchilov and Hutter(2017)}]{loshchilov2017decoupled}
Ilya Loshchilov and Frank Hutter. 2017.
\newblock Decoupled weight decay regularization.
\newblock \emph{arXiv preprint arXiv:1711.05101}.

\bibitem[{Lozhkov et~al.(2024)Lozhkov, Li, Allal, Cassano, Lamy{-}Poirier, Tazi, Tang, Pykhtar, Liu, Wei, Liu, Tian, Kocetkov, Zucker, Belkada, Wang, Liu, Abulkhanov, Paul, Li, Li, Risdal, Li, Zhu, Zhuo, Zheltonozhskii, Dade, Yu, Krau{\ss}, Jain, Su, He, Dey, Abati, Chai, Muennighoff, Tang, Oblokulov, Akiki, Marone, Mou, Mishra, Gu, Hui, Dao, Zebaze, Dehaene, Patry, Xu, McAuley, Hu, Scholak, Paquet, Robinson, Anderson, Chapados, and et~al.}]{starcoder2-24}
Anton Lozhkov, Raymond Li, Loubna~Ben Allal, Federico Cassano, Joel Lamy{-}Poirier, Nouamane Tazi, Ao~Tang, Dmytro Pykhtar, Jiawei Liu, Yuxiang Wei, Tianyang Liu, Max Tian, Denis Kocetkov, Arthur Zucker, Younes Belkada, Zijian Wang, Qian Liu, Dmitry Abulkhanov, Indraneil Paul, Zhuang Li, Wen{-}Ding Li, Megan Risdal, Jia Li, Jian Zhu, Terry~Yue Zhuo, Evgenii Zheltonozhskii, Nii Osae~Osae Dade, Wenhao Yu, Lucas Krau{\ss}, Naman Jain, Yixuan Su, Xuanli He, Manan Dey, Edoardo Abati, Yekun Chai, Niklas Muennighoff, Xiangru Tang, Muhtasham Oblokulov, Christopher Akiki, Marc Marone, Chenghao Mou, Mayank Mishra, Alex Gu, Binyuan Hui, Tri Dao, Armel Zebaze, Olivier Dehaene, Nicolas Patry, Canwen Xu, Julian~J. McAuley, Han Hu, Torsten Scholak, S{\'{e}}bastien Paquet, Jennifer Robinson, Carolyn~Jane Anderson, Nicolas Chapados, and et~al. 2024.
\newblock \href {https://doi.org/10.48550/ARXIV.2402.19173} {Starcoder 2 and the stack v2: The next generation}.
\newblock \emph{CoRR}, abs/2402.19173.

\bibitem[{Nishino et~al.(2019)Nishino, Takase, Hirao, and Nagata}]{nishino-etal-2019-generating}
Masaaki Nishino, Sho Takase, Tsutomu Hirao, and Masaaki Nagata. 2019.
\newblock \href {https://doi.org/10.18653/v1/D19-1674} {Generating natural anagrams: Towards language generation under hard combinatorial constraints}.
\newblock In \emph{Proceedings of the 2019 Conference on Empirical Methods in Natural Language Processing and the 9th International Joint Conference on Natural Language Processing (EMNLP-IJCNLP)}, pages 6408--6412, Hong Kong, China. Association for Computational Linguistics.

\bibitem[{OpenAI(2023)}]{gpt4}
OpenAI. 2023.
\newblock \href {https://doi.org/10.48550/ARXIV.2303.08774} {{GPT-4} technical report}.
\newblock \emph{CoRR}, abs/2303.08774.

\bibitem[{Peng et~al.(2024)Peng, Chai, and Li}]{peng-etal-2024-humaneval-xl}
Qiwei Peng, Yekun Chai, and Xuhong Li. 2024.
\newblock \href {https://aclanthology.org/2024.lrec-main.735} {{H}uman{E}val-{XL}: A multilingual code generation benchmark for cross-lingual natural language generalization}.
\newblock In \emph{Proceedings of the 2024 Joint International Conference on Computational Linguistics, Language Resources and Evaluation (LREC-COLING 2024)}, pages 8383--8394, Torino, Italia. ELRA and ICCL.

\bibitem[{Peng et~al.(2023)Peng, Weir, and Weeds}]{peng-etal-2023-testing}
Qiwei Peng, David Weir, and Julie Weeds. 2023.
\newblock \href {https://doi.org/10.18653/v1/2023.starsem-1.24} {Testing paraphrase models on recognising sentence pairs at different degrees of semantic overlap}.
\newblock In \emph{Proceedings of the 12th Joint Conference on Lexical and Computational Semantics (*SEM 2023)}, pages 259--269, Toronto, Canada. Association for Computational Linguistics.

\bibitem[{Provilkov et~al.(2020)Provilkov, Emelianenko, and Voita}]{provilkov-etal-2020-bpe}
Ivan Provilkov, Dmitrii Emelianenko, and Elena Voita. 2020.
\newblock \href {https://doi.org/10.18653/v1/2020.acl-main.170} {{BPE}-dropout: Simple and effective subword regularization}.
\newblock In \emph{Proceedings of the 58th Annual Meeting of the Association for Computational Linguistics}, pages 1882--1892, Online. Association for Computational Linguistics.

\bibitem[{Radford et~al.(2018)Radford, Narasimhan, Salimans, Sutskever et~al.}]{radford2018improving}
Alec Radford, Karthik Narasimhan, Tim Salimans, Ilya Sutskever, et~al. 2018.
\newblock Improving language understanding by generative pre-training.

\bibitem[{Rust et~al.(2023)Rust, Lotz, Bugliarello, Salesky, de~Lhoneux, and Elliott}]{pixel23}
Phillip Rust, Jonas~F. Lotz, Emanuele Bugliarello, Elizabeth Salesky, Miryam de~Lhoneux, and Desmond Elliott. 2023.
\newblock \href {https://openreview.net/pdf?id=FkSp8VW8RjH} {Language modelling with pixels}.
\newblock In \emph{The Eleventh International Conference on Learning Representations, {ICLR} 2023, Kigali, Rwanda, May 1-5, 2023}. OpenReview.net.

\bibitem[{Schuster and Nakajima(2012)}]{DBLP:conf/icassp/SchusterN12}
Mike Schuster and Kaisuke Nakajima. 2012.
\newblock \href {https://doi.org/10.1109/ICASSP.2012.6289079} {Japanese and korean voice search}.
\newblock In \emph{2012 {IEEE} International Conference on Acoustics, Speech and Signal Processing, {ICASSP} 2012, Kyoto, Japan, March 25-30, 2012}, pages 5149--5152. {IEEE}.

\bibitem[{Sennrich et~al.(2016)Sennrich, Haddow, and Birch}]{bpe}
Rico Sennrich, Barry Haddow, and Alexandra Birch. 2016.
\newblock \href {https://doi.org/10.18653/V1/P16-1162} {Neural machine translation of rare words with subword units}.
\newblock In \emph{Proceedings of the 54th Annual Meeting of the Association for Computational Linguistics, {ACL} 2016, August 7-12, 2016, Berlin, Germany, Volume 1: Long Papers}. The Association for Computer Linguistics.

\bibitem[{Sinha et~al.(2021)Sinha, Parthasarathi, Pineau, and Williams}]{sinha-etal-2021-unnatural}
Koustuv Sinha, Prasanna Parthasarathi, Joelle Pineau, and Adina Williams. 2021.
\newblock \href {https://doi.org/10.18653/v1/2021.acl-long.569} {{UnNatural} {L}anguage {I}nference}.
\newblock In \emph{Proceedings of the 59th Annual Meeting of the Association for Computational Linguistics and the 11th International Joint Conference on Natural Language Processing (Volume 1: Long Papers)}, pages 7329--7346, Online. Association for Computational Linguistics.

\bibitem[{Srivastava et~al.(2022)Srivastava, Rastogi, Rao, Shoeb, Abid, Fisch, Brown, Santoro, Gupta, Garriga{-}Alonso, Kluska, Lewkowycz, Agarwal, Power, Ray, Warstadt, Kocurek, Safaya, Tazarv, Xiang, Parrish, Nie, Hussain, Askell, Dsouza, Rahane, Iyer, Andreassen, Santilli, Stuhlm{\"{u}}ller, Dai, La, Lampinen, Zou, Jiang, Chen, Vuong, Gupta, Gottardi, Norelli, Venkatesh, Gholamidavoodi, Tabassum, Menezes, Kirubarajan, Mullokandov, Sabharwal, Herrick, Efrat, Erdem, Karakas, and et~al.}]{bigbench22}
Aarohi Srivastava, Abhinav Rastogi, Abhishek Rao, Abu Awal~Md Shoeb, Abubakar Abid, Adam Fisch, Adam~R. Brown, Adam Santoro, Aditya Gupta, Adri{\`{a}} Garriga{-}Alonso, Agnieszka Kluska, Aitor Lewkowycz, Akshat Agarwal, Alethea Power, Alex Ray, Alex Warstadt, Alexander~W. Kocurek, Ali Safaya, Ali Tazarv, Alice Xiang, Alicia Parrish, Allen Nie, Aman Hussain, Amanda Askell, Amanda Dsouza, Ameet Rahane, Anantharaman~S. Iyer, Anders Andreassen, Andrea Santilli, Andreas Stuhlm{\"{u}}ller, Andrew~M. Dai, Andrew La, Andrew~K. Lampinen, Andy Zou, Angela Jiang, Angelica Chen, Anh Vuong, Animesh Gupta, Anna Gottardi, Antonio Norelli, Anu Venkatesh, Arash Gholamidavoodi, Arfa Tabassum, Arul Menezes, Arun Kirubarajan, Asher Mullokandov, Ashish Sabharwal, Austin Herrick, Avia Efrat, Aykut Erdem, Ayla Karakas, and et~al. 2022.
\newblock \href {https://doi.org/10.48550/ARXIV.2206.04615} {Beyond the imitation game: Quantifying and extrapolating the capabilities of language models}.
\newblock \emph{CoRR}, abs/2206.04615.

\bibitem[{Sutskever et~al.(2011)Sutskever, Martens, and Hinton}]{10.5555/3104482.3104610}
Ilya Sutskever, James Martens, and Geoffrey Hinton. 2011.
\newblock Generating text with recurrent neural networks.
\newblock In \emph{Proceedings of the 28th International Conference on International Conference on Machine Learning}, ICML'11, page 1017–1024, Madison, WI, USA. Omnipress.

\bibitem[{Touvron et~al.(2023)Touvron, Martin, Stone, Albert, Almahairi, Babaei, Bashlykov, Batra, Bhargava, Bhosale, Bikel, Blecher, Canton{-}Ferrer, Chen, Cucurull, Esiobu, Fernandes, Fu, Fu, Fuller, Gao, Goswami, Goyal, Hartshorn, Hosseini, Hou, Inan, Kardas, Kerkez, Khabsa, Kloumann, Korenev, Koura, Lachaux, Lavril, Lee, Liskovich, Lu, Mao, Martinet, Mihaylov, Mishra, Molybog, Nie, Poulton, Reizenstein, Rungta, Saladi, Schelten, Silva, Smith, Subramanian, Tan, Tang, Taylor, Williams, Kuan, Xu, Yan, Zarov, Zhang, Fan, Kambadur, Narang, Rodriguez, Stojnic, Edunov, and Scialom}]{llama2}
Hugo Touvron, Louis Martin, Kevin Stone, Peter Albert, Amjad Almahairi, Yasmine Babaei, Nikolay Bashlykov, Soumya Batra, Prajjwal Bhargava, Shruti Bhosale, Dan Bikel, Lukas Blecher, Cristian Canton{-}Ferrer, Moya Chen, Guillem Cucurull, David Esiobu, Jude Fernandes, Jeremy Fu, Wenyin Fu, Brian Fuller, Cynthia Gao, Vedanuj Goswami, Naman Goyal, Anthony Hartshorn, Saghar Hosseini, Rui Hou, Hakan Inan, Marcin Kardas, Viktor Kerkez, Madian Khabsa, Isabel Kloumann, Artem Korenev, Punit~Singh Koura, Marie{-}Anne Lachaux, Thibaut Lavril, Jenya Lee, Diana Liskovich, Yinghai Lu, Yuning Mao, Xavier Martinet, Todor Mihaylov, Pushkar Mishra, Igor Molybog, Yixin Nie, Andrew Poulton, Jeremy Reizenstein, Rashi Rungta, Kalyan Saladi, Alan Schelten, Ruan Silva, Eric~Michael Smith, Ranjan Subramanian, Xiaoqing~Ellen Tan, Binh Tang, Ross Taylor, Adina Williams, Jian~Xiang Kuan, Puxin Xu, Zheng Yan, Iliyan Zarov, Yuchen Zhang, Angela Fan, Melanie Kambadur, Sharan Narang, Aur{\'{e}}lien Rodriguez, Robert Stojnic, Sergey Edunov,
  and Thomas Scialom. 2023.
\newblock \href {https://doi.org/10.48550/ARXIV.2307.09288} {Llama 2: Open foundation and fine-tuned chat models}.
\newblock \emph{CoRR}, abs/2307.09288.

\bibitem[{Xue et~al.(2022)Xue, Barua, Constant, Al{-}Rfou, Narang, Kale, Roberts, and Raffel}]{byt5}
Linting Xue, Aditya Barua, Noah Constant, Rami Al{-}Rfou, Sharan Narang, Mihir Kale, Adam Roberts, and Colin Raffel. 2022.
\newblock \href {https://doi.org/10.1162/TACL\_A\_00461} {Byt5: Towards a token-free future with pre-trained byte-to-byte models}.
\newblock \emph{Trans. Assoc. Comput. Linguistics}, 10:291--306.

\end{thebibliography}

\appendix

\clearpage
% \section{Appendix}
% \label{sec:appendix}
%%%%%%%%%%%%%%%%%%%%%%%%%%%%%%%%
% \onecolumn
% \tableofcontents
% \twocolumn
%%%%%%%%%%%%%%%%%%%%%%%%%%%%%%%%

\section{Task Examples}
\label{ap:example}
\subsection{Complex Problem Solving Examples}
\label{ap:example-complex}
 
% \paragraph{Anagram Task}
We present detailed examples of complex problem-solving tasks including word anagram and identifying math theorems as follows:

% colbacktitle=white!80!gray, coltitle=black, 
\begin{tcolorbox}[lowerbox=invisible, colback=white, fonttitle = \bfseries, fontupper = \sffamily, fontlower = \itshape, title=Anagram Task Format]
\textbf{Input}: A string of jumbled 
  characters (\textit{e.g.}, ``moeh'' for ``home''). \\
\textbf{Output}: The correct unscrambled word (\textit{e.g.}, ``home'').
\end{tcolorbox}

% \paragraph{}

% The dataset comprises 54 problems divided into nine sections, each representing a major area of mathematics research or advanced pedagogy. The topics include Elementary Number Theory, Calculus, Real Analysis, Group Theory, Geometry, Theoretical Computer Science, Linear Algebra, Topology, and Statistics. Each problem is designed to compile correctly in a base {\LaTeX} compiler, ensuring the model must tokenize and interpret structured mathematical input precisely. 

\begin{tcolorbox}[lowerbox=invisible, colback=white, fonttitle = \bfseries, fontupper = \sffamily, fontlower = \itshape, title=Identifying Math Theorems Task Format]
\textbf{Input}: A {\LaTeX}-formatted mathematical theorem. \textit{E.g.}, ``Let $f\in L^1(\mathbb{R})$ be an integrable function. The span of $\{f_a(x) = f(x + a):a\in\mathbb{R}\}$ is dense in $L^1(\mathbb{R})$ if and only if $\widehat{f}$ has no real roots..
\begin{enumerate}[label=\Alph*), noitemsep, left=0pt, labelsep=4pt, topsep=0pt, partopsep=0pt]
    \item Let $f\in L^1(\mathbb{R})$ be an integrable function. The span of $\{f_a(x) = f(x + a):a\in\mathbb{R}\}$ is dense in $L^1(\mathbb{R})$ if and only if $\widehat{f}$ has no real roots. 
     \item Let $f\in L^1(\mathbb{R})$ be an integrable function. The span of $\{f_a(x) = f(x + a):a\in\mathbb{R}\}$ is dense in $L^1(\mathbb{R})$ if and only if $\widehat{f}$ has no real roots. .
    \item Let $f\in L^1(\mathbb{R})$ be an integrable function. The span of $\{f_a(x) = f(x + a):a\in\mathbb{R}\}$ is dense in $L^1(\mathbb{R})$ if and only if $\widehat{f}$ is irreducible over $\mathbb{Q}$.
    \item Let $f\in L^1(\mathbb{R})$ be an integrable function. The span of $\{f_a(x) = f(x + a):a\in\mathbb{R}\}$ is dense in $L^1(\mathbb{R})$ if and only if $\widehat{f}$ has no repeated roots.
\end{enumerate}
\textbf{Output}: The model must determine whether the theorem is true. If it is false, the model should provide the correct version; \emph{i.e.}, select the option ``A''.
\end{tcolorbox}

\subsection{Intra-Token Probing Examples}
\label{ap:example-intra-token}
For intra-token probing tasks, we provide Character Count (CC), $N$-th Character (NC), $N$-th Character Reverse (NCR), and Case Conversion (CCV) for illustration.

% \vspace{0.4em}
% left=0pt, right=0pt, top=2pt, bottom=1pt, before skip=1pt, after skip=1pt
\begin{tcolorbox}[lowerbox=invisible, colback=white, fonttitle = \bfseries, fontupper = \sffamily, fontlower = \itshape, title=Character Count (CC) ]
\textbf{Input}: Which character appears 3 times in the word `messrs'? \\
\textbf{Output}: `s'.
\end{tcolorbox}

\begin{tcolorbox}[lowerbox=invisible, colback=white, fonttitle = \bfseries, fontupper = \sffamily, fontlower = \itshape, title=$N$-th Character (NC)]
\textbf{Input}: What is the 4th character of the word `myron'? \\
\textbf{Output}: `o'.
\end{tcolorbox}

\begin{tcolorbox}[lowerbox=invisible, colback=white, fonttitle = \bfseries, fontupper = \sffamily, fontlower = \itshape, title=$N$-th Character Reverse (NCR)]
\textbf{Input}: What is the 2nd character from the end of the word `pensioner'? \\
\textbf{Output}: `e'.
\end{tcolorbox}

\begin{tcolorbox}[lowerbox=invisible, colback=white, fonttitle = \bfseries, fontupper = \sffamily, fontlower = \itshape, title=Case Conversion (CCV)]
\textbf{Input}: Which character appears 3 times in the word `messrs'? \\
\textbf{Output}: `s'.
\end{tcolorbox}

\subsection{Inter-Token Probing Examples}
\label{ap:example-inter-token}
We present examples of inter-token probing tasks, which involve identifying Common Substrings (CS), Longest Common Subsequences (LCSeq), and Longest Common Substrings (LCS). These tasks evaluate the model's ability to analyze and compare internal structure across different inputs.

\begin{tcolorbox}[lowerbox=invisible, colback=white, fonttitle = \bfseries, fontupper = \sffamily, fontlower = \itshape, title=Common Substrings (CS) ]
\textbf{Input}: What are the common substrings of 'critical' and 'conscious'? \\
\textbf{Output}: `i`, `c'.
\end{tcolorbox}

\begin{tcolorbox}[lowerbox=invisible, colback=white, fonttitle = \bfseries, fontupper = \sffamily, fontlower = \itshape, title=Longest Common Subsequences (LCSeq)]
\textbf{Input}: What are the longest common subsequences of `illustrate' and `critical'? \\
\textbf{Output}: `ita'.
\end{tcolorbox}

\begin{tcolorbox}[lowerbox=invisible, colback=white, fonttitle = \bfseries, fontupper = \sffamily, fontlower = \itshape, title=Longest Common Substrings (LCS)]
\textbf{Input}: What are the longest common substrings of `cow' and `condition'? \\
\textbf{Output}: `co'.
\end{tcolorbox}

\section{Experimental Settings}
\label{ap:settings}

\subsection{Baselines}
We include Llama3-8B, Llama3-8B-Instruct, Llama3-70B, Mistral-7B and Mixtral-8x7B for LLM evaluation.
\paragraph{Llama3}
Llama3~\cite{llama3modelcard} series are one of the most powerful open-sourced models recently. Llama3-8B is a dense pretrained model with a vocab size of 128256, which needs few-shot examples to better follow instructions. Llama3-8B-Instruct is also envolved for diverse model types. Llama3-8B-Instrcut is a instruction-fine-tuned version of Llama3-8B, showing much improvement over Llama3-8B on benchmarks like HumanEval and TruthfulQA. 

\paragraph{Mistral \& Mixtral}
Mistral-7B~\cite{mistral23} is a dense model with a vocab size of 32000 released last year. Mixtral-8x7B~\cite{mixtral24} is a sparse mixture-of-expert(MoE) model with 13B active parameters, whose performance greatly surpasses Mistral-7B and matches the performance of Llama2-70B.

\paragraph{GPT-4 Turbo}
The model version we evaluate is \texttt{``gpt-4-1106-preview''}\footnote{\url{https://platform.openai.com/docs/models/gpt-4-turbo-and-gpt-4}}. Compared with GPT-4, "gpt-4-1106-preview" yields stronger performance on following instructions, structured output and other abilities.

\subsection{Evaluation Settings}
We utilize lm-evaluation-harness~\cite{eval-harness} for the evaluation of GSM8K, MMLU, and TruthfulQA. For HumanEval, we adopt bigcode-evaluation-harness~\cite{bigcode-evaluation-harness}. All models are tested under bfloat16 precision for higher efficiency.

To eliminate the impact of the Chain-of-Thought (CoT) prompt, GSM8K is evaluated using a 5-shot setting without CoT, and we report the "exact match" as the final metric. For HumanEval, we report pass@1 using a temperature of 0.2 and a top-$p$ of 0.95. The maximum total length of the prompt and model output is set to 512 for Llama3 and 1024 for Mistral-7B and Mixtral-8x7B. We only apply corruption to the annotation to make sure that entry point can be found after corruption. For TruthfulQA, model performances are measured within the ``MC1 (single-true)'' setting.

For models after instruction-tuning, we do not apply chat templates except for TruthfulQA, as model outputs tend to be difficult to parse in chat mode.

\subsection{Post-Training Details}
\label{ap:train_details}
\paragraph{Dataset}
We synthesized the dataset for RQ2 with a template-based method. Table~\ref{tab:app-ft-dataset} describes statistics of dataset splits for each task.
\begin{table}[!ht]
\centering
\resizebox{\columnwidth}{!}{
\ttfamily
\begin{tabular}{lrr}
\\ \toprule
\textbf{Task}  & \textbf{Train} & \textbf{Test} \\ \midrule
Intra-Token Probing & 111,070 & 800 \\ \midrule
\hspace{1em}Character Count   &  20,775  & 200   \\
\hspace{1em}N-th Character   &  31,241   & 200   \\
\hspace{1em}N-th Character Reverse & 31,316 & 200 \\
\hspace{1em}Case Conversion & 27,738 & 200 \\ \midrule
Inter-Token Probing & 14,400 & 600 \\ \midrule
\hspace{1em}Common Substrings & 4,800 & 200 \\
\hspace{1em}Longest Common Substrings & 4,800 & 200 \\
\hspace{1em}Longest Common Subsequences & 4,800 & 200 \\\bottomrule
\end{tabular}
}
\caption{The dataset size for training and testing.}
\label{tab:app-ft-dataset}
\end{table}

% \paragraph{Training Details}
% We employ the AdamW optimizer~\cite{loshchilov2017decoupled} with the hyperpameters of $\beta_1=0.9$, $\beta_2=0.95$. The peak learning rate is set to 5e-5, and the minimum learning rate is set to 1e-6. The learning rate warms up during the first 10\% of training steps and then decays with a cosine scheduler.

% Given the difference in data distribution resulting from BPE-dropout, we post-train Mistral-7B~\cite{mistral23} on the training split for 5 epochs with a global batch size of 16. Following the pre-training recipe, we concatenate all sequences and then chunk them into fixed context lengths of 4096 for our autoregressive post-training. We conduct data shuffling within the same epochs.

% Since tokenization results can vary for the same sentence after BPE-Dropout, we do not shuffle data between epochs to maintain consistent training distribution.

\section{Details of Probing Task Construction}
\label{ap:task_construction}

\subsection{Token Structure Probing (RQ2)}
To create a comprehensive test set for evaluating the tokenization capabilities of LLMs, we followed a systematic data synthesis process. Initially, we manually collected a set of around 300 words from the web, ensuring a diverse representation of word structures. This collection included words with common suffixes, prefixes, and varying lengths to cover a broad range of token structures.

Next, we defined a set of rules to create tasks for both intra-token and inter-token evaluations. These rules were designed to test different aspects of tokenization, such as character counting, character identification, case conversion, and identifying common substrings and subsequences. 

Then we generated the probing tasks. For intra-token evaluations, we created tasks like Character Count (CC), $N$-th Character (NC), $N$-th Character Reverse (NCR), and Case Conversion (CCV). For inter-token evaluations, we developed tasks such as Common Substrings (CS), Longest Common Substrings (LCS), and Longest Common Subsequences (LCSeq). Each task was carefully crafted to test the model’s ability to understand and manipulate token structures at various levels.

\subsection{Typographical Variation Task (RQ3)}

\begin{table}[!ht]
\centering
{
\ttfamily
\begin{tabular}{lr}
\\ \toprule
\textbf{Task} & \textbf{Test} \\ \midrule
GSM8k & 1,319 \\ 
MMLU  & 14,042   \\
TruthfulQA & 817   \\
HumanEval & 164 \\ \bottomrule
\end{tabular}
}
\caption{Dataset statistics of typographical variation tasks (RQ3).}
\label{tab:app-rq3-dataset}
\end{table}
To thoroughly evaluate the typographical variation (RQ3), we conduct corruption to the questions of several benchmark datasets' test split and keep answers intact. Details of evaluation dataset are summarized at Table~\ref{tab:app-rq3-dataset}.

\begin{figure*}[!ht]
    \centering
    \includegraphics[width=0.95\linewidth]{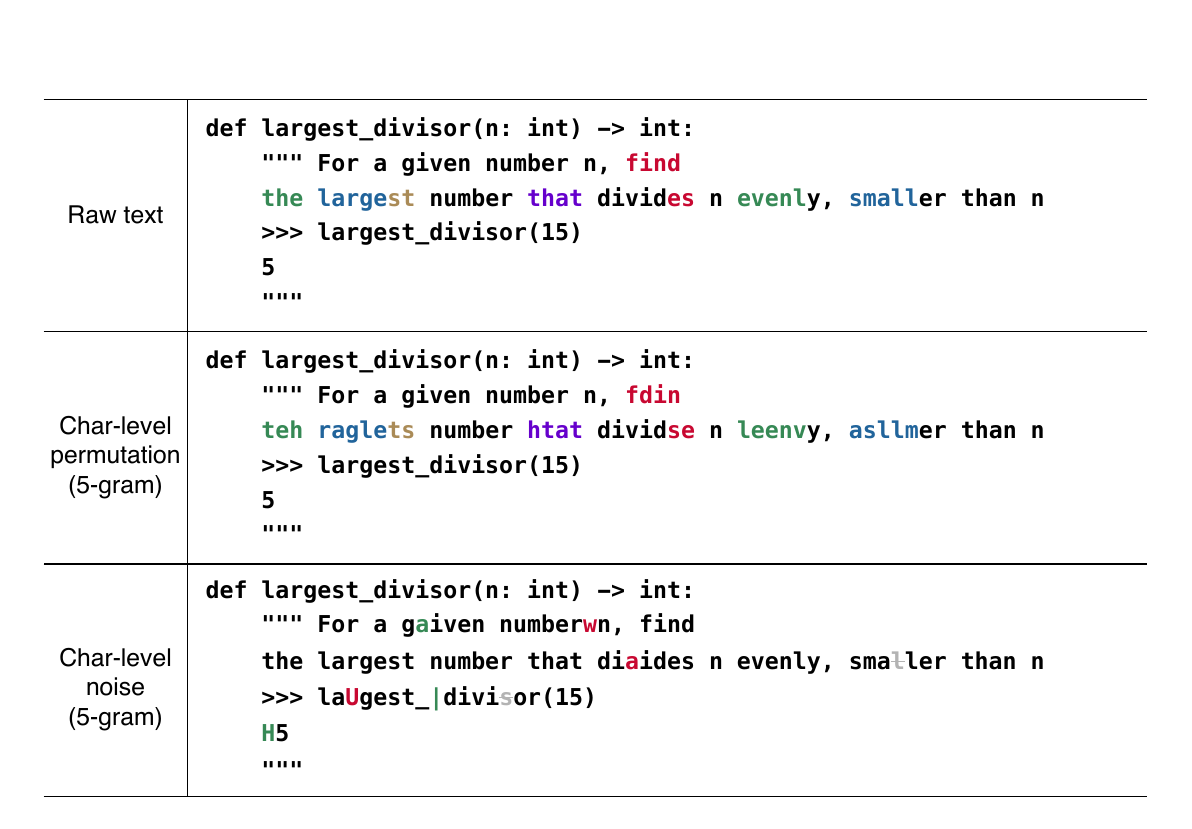}
        \vspace{-10pt}
    \caption{Example of a char-level(5-gram) corrupted prompt from HumanEval. \textcolor[RGB]{200,8,49}{Red}, \textcolor[RGB]{54,137,85}{Green}, and \textcolor[RGB]{179,179,179}{Gray} denote replacement, insertion, and deletion respectively.}
    \label{fig:app-case-char} 
\end{figure*}

\begin{figure*}[!ht]
    \centering
    \includegraphics[width=0.95\linewidth]{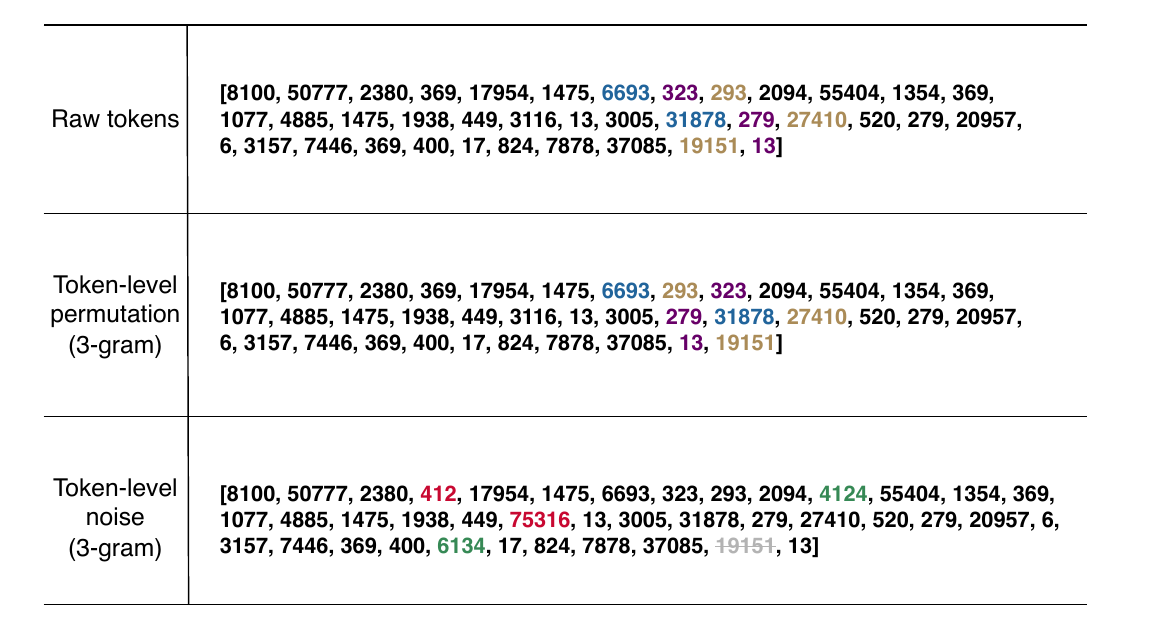}
    \vspace{-10pt}
    \caption{Example of a token-level(tri-gram) corrupted prompt from GSM8K. \textcolor[RGB]{200,8,49}{Red}, \textcolor[RGB]{54,137,85}{Green}, and \textcolor[RGB]{179,179,179}{Gray} denote replacement, insertion, and deletion respectively.}
    \label{fig:app-case-token}
        \vspace{-10pt}
\end{figure*}

\paragraph{Permutation}
For permutation task, we randomly shuffle tokens or characters within an $n$-gram range with a 50\% probability. We evaluate by using $n$-gram range from 2 to 5, in which various $n$-gram levels could assess the model's performance under varying degrees of corruption.

\paragraph{Noise Injection}
We consider three kinds of noise that commonly encountered by humans: insertion (10\%), deletion (10\%), and replacement (10\%). For insertion, we choose a token or character from the current $n$-gram under 50\% circumstances. Otherwise, we randomly select a token or character from the whole vocabulary (token or character) and add it to a random position.

When performing replacing, we replace a token or character in the current $n$-gram with a token or character randomly selected from the whole vocabulary. 

\paragraph{Character-Level}
We performed character-level permutation within word boundaries, ensuring that the positions of punctuation marks or spaces remained unchanged after permutation. Even when there were not enough characters to form a complete $n$-gram, permutation was still applied. We ignore word boundaries when injecting character-level noise. Figure~\ref{fig:app-case-char} illustrates our character-level corruption approach using a 5-gram granularity.

\paragraph{Token-Level}
We converted the input prompts into tokens using tokenizers from different model families. Corruption was then applied within an $n$-gram range, without regard to word boundaries. This approach simulates token-level typographical variations, challenging the models to handle disruptions at the token level. Figure~\ref{fig:app-case-token} gives an example of how we apply such variations to an encoded prompt.

\section{Detailed Results of BPE-dropout Post-Training}
\label{ap:results}

% \begin{figure*}[!ht] 
% \centering
% % \vspace{-1mm}
% \subfigure[0-shot] { \label{fig:epoch_0_shot}
% \includegraphics[width=0.48\linewidth]{figure/ft_epoch_0_shot.pdf}
% }
% % \vspace{-1mm}
% \hfill
% \subfigure[1-shot] { \label{fig:epoch_1_shot}
% \includegraphics[width=0.48\linewidth]{figure/ft_epoch_1_shot.pdf}
% }
% \subfigure[2-shot] { \label{fig:2-shot}
% \includegraphics[width=0.48\linewidth]{figure/ft_epoch_2_shot.pdf}
% }
% \subfigure[3-shot] { \label{fig:3-shot}
% \includegraphics[width=0.48\linewidth]{figure/ft_epoch_3_shot.pdf}
% }
% % \vspace{-1mm}
% \caption{The impact of BPE-dropout on EM scores across seven tasks (CS, LCSeq, CCV, CC, NC, LCS, NCR) under different post-training conditions: (a) 0-shot, (b) 1-shot, (c) 2-shot, and (d) 3-shot. The dropout rates range from 0.0 to 0.8. The plots show that moderate dropout rates generally lead to improvements. Tasks such as CS and CC are more robust to dropout, maintaining higher scores even at moderate dropout rates, while tasks like NC, LCS, and NCR show significant performance drops with increasing dropout.}
% \label{fig:ft_epoch}
% % \vspace{-1mm}
% \end{figure*}

The results, as shown in Figure~\ref{fig:ft_epoch}, illustrate the effect of BPE-dropout on EM scores across seven tasks (CS, LCSeq, CCV, CC, NC, LCS, NCR) under various post-training conditions (0-shot, 1-shot, 2-shot, and 3-shot) with dropout rates ranging from 0.0 to 0.8.

We observe that moderate dropout rates (0.2 to 0.4) appear to improve the convergence of EM scores across all tasks, particularly in the zero-shot setting. This indicates that a certain level of variability introduced by dropout can help the model generalize better when no additional examples are provided. In the 1-shot, 2-shot, and 3-shot settings, moderate dropout rates contribute to stabilizing performance, suggesting that this level of dropout introduces useful regularization without significantly compromising token integrity.

It is evident that higher dropout rates (0.6 and 0.8) lead to a noticeable decline in EM scores across all tasks and post-training conditions. This indicates that excessive dropout disrupts the tokenization process, resulting in subwords with fewer merges that conflict with the original pre-training of the models. Tasks such as NC, LCS, and NCR show more significant drops in EM scores with higher dropout rates, reflecting their complexity and the challenge of maintaining token integrity under substantial dropout.

Task-specific performance varies under different dropout conditions. CS and CC tasks exhibit high robustness to dropout, maintaining relatively stable EM scores even at moderate dropout rates. This suggests that the nature of these tasks makes them less sensitive to the variability introduced by dropout. LCSeq and CCV tasks show moderate sensitivity to dropout, with a noticeable decline in performance at higher dropout rates. NC, LCS, and NCR tasks are more adversely affected by higher dropout rates, indicating their reliance on stable token sequences.

The positive effects of moderate BPE-dropout include improved generalization and regularization. Moderate dropout rates (0.2 to 0.4) introduce beneficial variability, enhancing the model's ability to generalize, particularly in zero-shot scenarios where the model must rely solely on its pre-trained knowledge without additional examples. BPE-dropout acts as a regularizer, preventing the model from overfitting to specific token sequences seen during pre-training. This is especially useful in low-resource settings (e.g., 0-shot and 1-shot) where overfitting can be a significant concern.

However, challenges arise with high dropout rates. Higher dropout rates lead to subwords with fewer merges, deviating from the token sequences the model encountered during pre-training. This disruption results in poorer performance, as the model struggles to reconcile the altered tokenization with its pre-trained representations. More complex tasks (NC, LCS, NCR) suffer more from high dropout rates, highlighting the need for careful tuning of dropout rates based on task complexity and tokenization stability requirements.

The findings suggest that there is an optimal range for BPE-dropout rates that balances the benefits of regularization and improved generalization with the need to maintain token integrity. Practitioners should consider moderate dropout rates to leverage the positive effects while avoiding the pitfalls of excessive dropout. Different tasks exhibit varying sensitivities to dropout, underscoring the importance of task-specific dropout tuning to achieve the best performance outcomes.

% \subsection{Experiments on Spelling Bee}
% We further examine Llama3 models on BIG-Bench Spellling Bee task. The results are shown in Table 

% \subsection{Additional Experiments}
% Spelling Bee @pqw

\section{Additional Analysis}
\label{ap:analysis}

\subsection{Impact of Typographical Variations on Sequence Length}
\begin{figure}[!ht]
    \vspace{-1mm}
    \centering
    \includegraphics[width=0.95\linewidth]{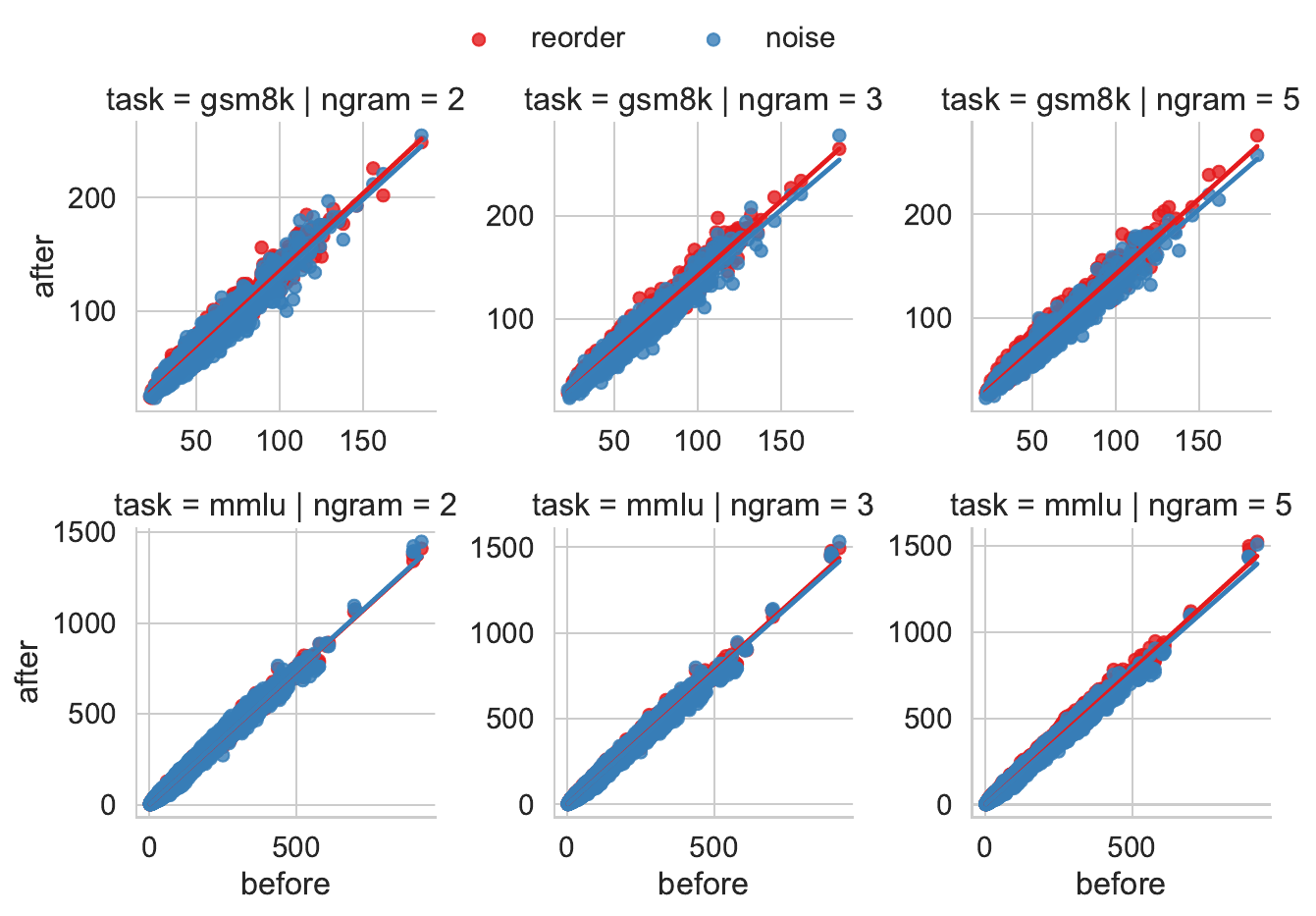}
     % \vspace{-1em}
    \caption{Scatter plots showing the positive correlation between token lengths before and after introducing typographical errors on MMLU and GSM8K, varying $n$-gram settings (2, 3, or 5). The x-axis represents the original token length, and the y-axis represents token length after adding errors.
    }
    \label{fig:typo_len_analysis} 
    \vspace{-1em}
\end{figure}

Figure~\ref{fig:typo_len_analysis} shows a strong positive correlation between token lengths before and after introducing typographical errors across different tasks (GSM8K and MMLU) and $n$-gram settings (2, 3, 5). We observe that token lengths after introducing errors are proportional to their original lengths.
Besides, both GSM8k and MMLU tasks exhibit similar patterns, and the $n$-gram settings (2, 3, 5) do not significantly alter this relationship. Most interestingly, the slope for reorder errors is relatively larger than for noise errors, indicating that \textbf{reorder errors tend to result in a slightly greater increase in token length compared to noise errors}.

\subsection{Compositional Challenges in Token Embeddings}
\label{ap:token_embed}
Figure~\ref{fig:vec_plot} illustrates the compositional challenges faced by existing LLMs when handling subword units. Specifically, it presents cosine similarities and angular differences between embeddings of original words and their subword components.

In Figure~\ref{fig:vec_assignment}, we observe the word ``assignment'' and its subword components ``assign'' and ``ment''. The cosine similarity between the composite embedding ``assign + ment'' and the original word "assignment" is relatively low at 0.21, with a significant angular difference of 78.16 degrees. This substantial disparity indicates that the learned token embeddings fail to capture the surface form composition accurately. The model does not recognize that ``assign'' combined with ``ment'' should semantically align closely with ``assignment''.

Similarly, Figure~\ref{fig:vec_import} shows the word ``import'' and its subword components ``im'' and ``port''. Here, the cosine similarity between ``im + port'' and ``import'' is 0.13,  and the angular difference is still notable at 82.47 degrees. This suggests that while the model captures some compositional aspects, it still struggles to fully integrate subword information to match the original word's embedding perfectly.

These observations highlight a critical limitation of existing LLMs: their learned token embeddings do not adequately capture the surface form composition. The inability to effectively combine subword units to represent the full word's meaning undermines the model's overall understanding and processing capabilities. 

% \paragraph{Discussion}

% \comm{+add input ppl/ed (noise/perm) vs. score regplot ? Corr.}

% \comm{For BPE merged tokens, observe embedding: A+B → AB (cos?)  }

% (Before/after finetuning?) -> x little change 

\end{document}